%% file: pdfbench.tex
\documentclass{article}

\usepackage[accepted]{icml2026}

\usepackage{amsmath}
\usepackage{amssymb}
\usepackage{mathtools}
\usepackage{amsthm}
\usepackage[utf8]{inputenc} % allow utf-8 input
\usepackage[T1]{fontenc}    % use 8-bit T1 fonts
\usepackage{hyperref}       % hyperlinks
\usepackage{url}            % simple URL typesetting
\usepackage{booktabs}       % professional-quality tables
\usepackage{amsfonts}       % blackboard math symbols
\usepackage{nicefrac}       % compact symbols for 1/2, etc.
\usepackage{microtype}      % microtypography
\usepackage{xcolor}         % colors
\usepackage{graphicx}
\usepackage{caption}
\usepackage{subcaption}
\usepackage{xspace}
\usepackage{tabularx}
\usepackage{algorithm}
\usepackage{algorithmic}
\usepackage{enumitem}
\usepackage{multirow}
\usepackage{makecell}
\usepackage{color,colortbl}
\usepackage[most]{tcolorbox}
\usepackage{pifont}
\usepackage[capitalize,noabbrev]{cleveref}

\theoremstyle{plain}

\theoremstyle{definition}

\theoremstyle{remark}

\usepackage[disable,textsize=tiny]{todonotes}
\definecolor{takeawaybg}{HTML}{F7F7F7}
\definecolor{takeawayframe}{HTML}{555555}
\newtcolorbox{takeawaybox}[1][]{
  enhanced,
  breakable,
  colback=takeawaybg,
  colframe=takeawayframe,
  boxrule=0.8pt,
  leftrule=2.0pt,
  arc=2pt,
  outer arc=2pt,
  boxsep=0pt,
  left=6pt,
  right=6pt,
  top=5pt,
  bottom=5pt,
  before skip=0.6\baselineskip,
  after skip=0.6\baselineskip,
  fontupper=\small,
  #1
}

\newcounter{takeawaycounter}
\newcommand{\takeaway}[1]{%
  \stepcounter{takeawaycounter}%
  \begin{takeawaybox}
    \textbf{Takeaway~\Roman{takeawaycounter}.}%
    \quad%
    \textit{#1}%
  \end{takeawaybox}
}

\definecolor{First}{HTML}{50aaf0}
\definecolor{Second}{HTML}{6BC3E7}
\definecolor{Third}{HTML}{ACEEEE}
\newcommand{\mnum}{16\xspace}
\newcommand{\first}{\cellcolor{First}}
\newcommand{\second}{\cellcolor{Second}}
\newcommand{\third}{\cellcolor{Third}}
\newcommand{\firsttext}{\textcolor{First}}
\newcommand{\secondtext}{\textcolor{Second}}
\newcommand{\thirdtext}{\textcolor{Third}}
\newcommand{\ourbench}{\textsc{PDFBench}\xspace}
\newcommand{\desctest}{\textit{MolinstTest}\xspace}
\newcommand{\keywtest}{\textit{SwissTest}\xspace}
\newcommand{\denovo}{\textit{de novo}\xspace}
\newcommand{\Denovo}{\textit{De novo}\xspace}
\newcommand{\cmark}{\textcolor{green}{\ding{51}}}
\newcommand{\xmark}{\textcolor{red}{\ding{55}}}

\icmltitlerunning{PDFBench: A Benchmark for \Denovo Protein Design from Function}

\begin{document}

\twocolumn[
  \icmltitle{PDFBench: A Benchmark for \Denovo Protein Design from Function}



  \icmlsetsymbol{equal}{*}

  \begin{icmlauthorlist}
    \icmlauthor{Jiahao Kuang}{ecnu,equal}
    \icmlauthor{Nuowei Liu}{ecnu,equal}
    \icmlauthor{Changzhi Sun}{teleai}
    \icmlauthor{Jie Wang}{ecnu}
    \icmlauthor{Tao Ji}{fudan}
    \icmlauthor{Yuanbin Wu}{ecnu}
  \end{icmlauthorlist}

  \icmlaffiliation{ecnu}{School of Computer Science and Technology, East China Normal University}
  \icmlaffiliation{teleai}{Institute of Artificial Intelligence (TeleAI), China Telecom}
  \icmlaffiliation{fudan}{College of Foreign Languages and Literatures, Fudan University}

  \icmlcorrespondingauthor{Tao Ji}{taoji@fudan.edu.cn}
  \icmlcorrespondingauthor{Yuanbin Wu}{ybwu@cs.ecnu.edu.cn}

  \icmlkeywords{De novo Protein Design, Evaluation Metric, Benchmark}

  \vskip 0.3in
]



\printAffiliationsAndNotice{\icmlEqualContribution}

\begin{abstract}
Function-guided protein design is a crucial task with significant applications in drug discovery and enzyme engineering. However, the field lacks a unified and comprehensive evaluation framework. Current models are assessed using inconsistent and limited subsets of metrics, which prevents fair comparison and a clear understanding of the relationships between different evaluation criteria. To address this gap, we introduce \ourbench, the first comprehensive benchmark for function-guided \denovo protein design. Our benchmark systematically evaluates eight state-of-the-art models on \mnum metrics across two key settings: description-guided design, for which we repurpose the Mol-Instructions dataset, originally lacking quantitative benchmarking, and keyword-guided design, for which we introduce a new test set, \keywtest, created with a strict datetime cutoff to ensure data integrity. By benchmarking across a wide array of metrics and analyzing their correlations, \ourbench enables more reliable model comparisons and provides key insights to guide future research. The codes and datasets are available in \url{https://github.com/PDFBench/PDFBench}.

\end{abstract}

\section{Introduction}
\label{sec:intro}
\input{docs/010Introduction}

\section{PDFBench} 
\label{sec:bench}
\input{docs/030Bench}

\section{Results}
\label{sec:res}
\input{docs/050Results}

\section{Rethinking the Evaluation Metrics}
\label{sec:discussion}
\input{docs/060Discussion}

\section{Related Work}
\label{sec:related}
\input{docs/020RelatedWork}

\section{Conclusion}
\label{sec:conclusion}
\input{docs/070Conclusion}

\section*{Impact Statement}

This paper presents work whose goal is to advance the field of Machine
Learning. There are many potential societal consequences of our work, none
which we feel must be specifically highlighted here.

\section*{Acknowledgements}
The authors wish to thank all reviewers for their helpful comments and suggestions. This work was partially funded by the National Natural Science Foundation of China (No.62506079).

\bibliography{pdfbench}
\bibliographystyle{icml2026}

% APPENDIX
\newpage
\appendix
\onecolumn
\input{docs/080Appendix}

\end{document}

%% file: docs/010Introduction.tex
Proteins are essential macromolecules that play a key role in many biological processes by performing a wide range of functions. Protein design \citep{huang2016coming,albanese2025computational} is of great significance in areas such as enzyme engineering \citep{planas2021computational,kries2013novo} and drug discovery \citep{hung2014computational,tiwari2022computational}. Compared with unconditional generation \citep{ferruz2022protgpt2}, generation guided by specific functions \citep{lee2024robust} or control tags \citep{hayes2025simulating}, protein design based on user-specified functions provides greater practical value \citep{liu2025text,madani2023large}.
Current protein design from function tasks can be categorized into two types, description-guided (which utilizes textual functional descriptions as input) and keyword-guided (which employs function keywords such as InterPro (IPR) \citep{hunter2009interpro} entries or Gene Ontology (GO) \citep{gene2021gene} terms as input).

\begin{figure*}[!htb]
    \centering
    \begin{subfigure}[b]{0.6\textwidth}
        \centering
        \includegraphics[width=\textwidth]{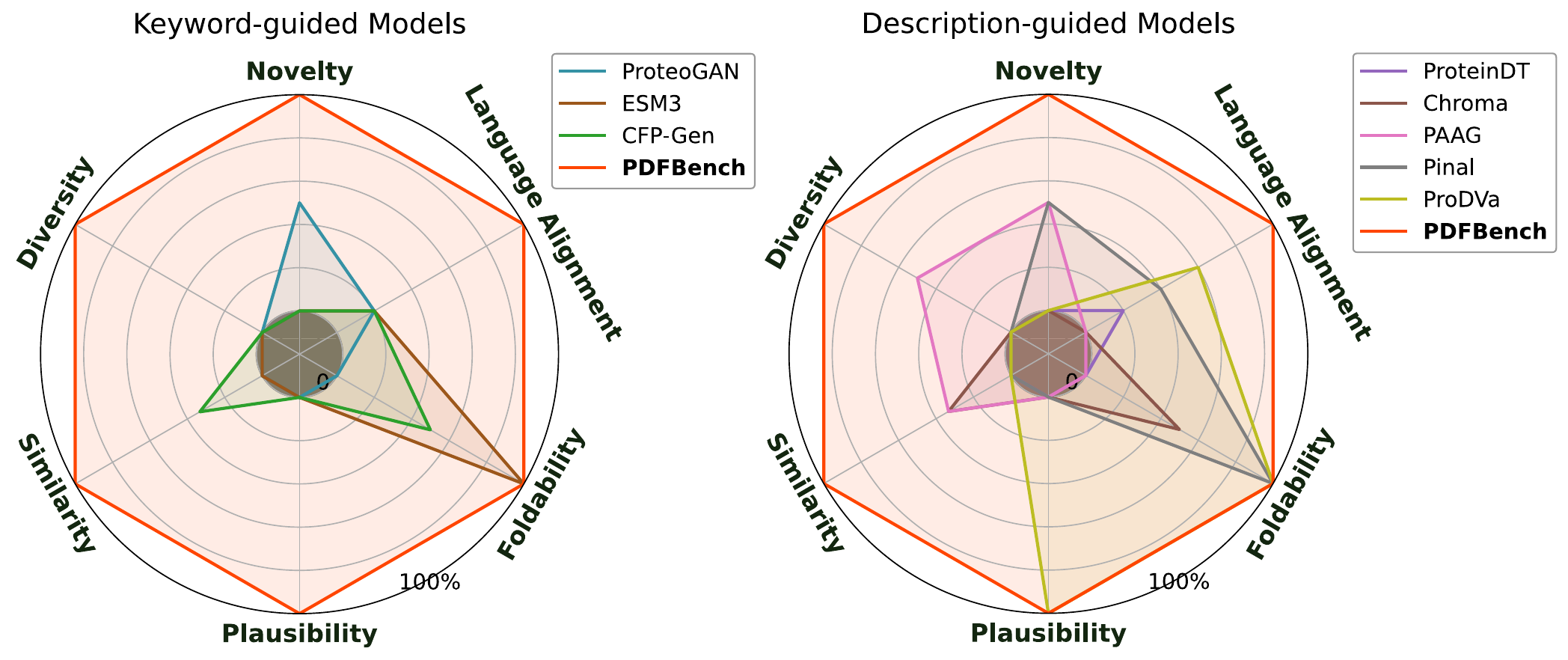}
        \caption{}
    \end{subfigure}\begin{subfigure}[b]{0.4\textwidth}
        \centering
        \includegraphics[width=\textwidth]{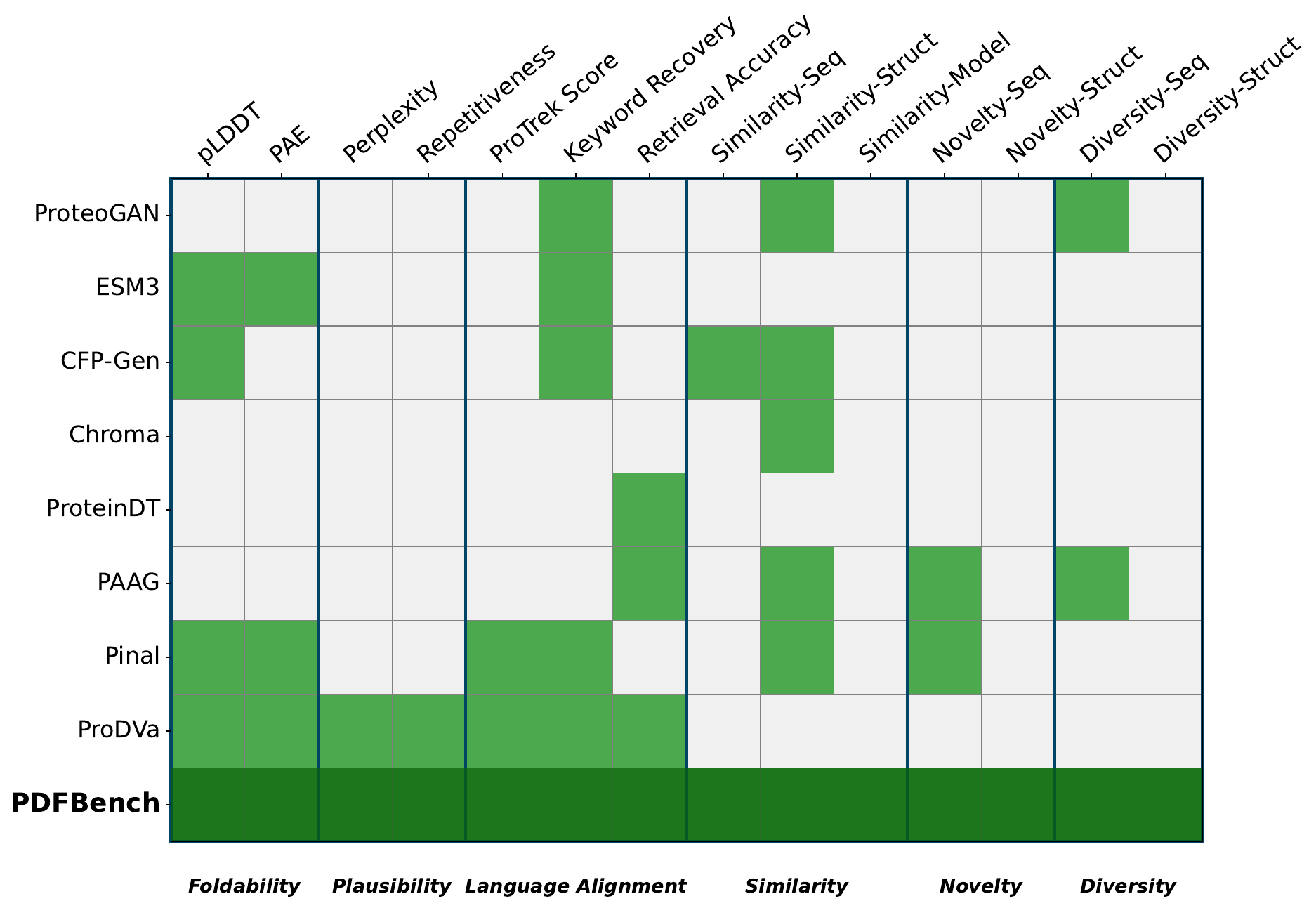}
        \caption{}
    \end{subfigure}
    \caption{Overview of current function-guided protein design models evaluated using different metrics, highlighting the lack of a unified and comprehensive evaluation framework. (a) Proportion of metrics employed in each previous work. In \ourbench, metrics are categorized into 6 dimensions, and we show that none of the prior works have been evaluated across all dimensions. (b) Detailed view of the metrics in \ourbench, with several representative metrics from each dimension presented.}
    \label{fig:intro-metrics}
\end{figure*}

To build protein design models, 
choosing proper evaluation metrics is crucial.
Various metrics (e.g., language alignment, foldability) are applied in the literature,
but different models are usually evaluated 
in different ways. 
%various evaluation metrics (e.g., language alignment, foldability) are applied.
In Figure~\ref{fig:intro-metrics}, 
we list existing evaluation metrics and depict how typical protein design models are evaluated against them. It shows that among the eight representative models, only half of them assess foldability, and only quarter of them evaluate novelty and diversity. For language alignment metrics, models either ignore model-based or retrieval-based metrics. Moreover, none of them consider all dimensions across the different metrics.

The lack of a comprehensive evaluation across all metrics can lead to unfair comparisons between methods. Furthermore, the correlations among different metrics have not been thoroughly investigated, limiting the understanding of these metrics and thereby hindering effective and insightful future research.

To bridge these gaps, we propose \ourbench, which, to the best of our knowledge, is the first benchmark designed to evaluate the capabilities of 8 novel function-guided \denovo protein design methods, encompassing both description-guided approaches and keyword-guided approaches. Comprehensive benchmarking of the description-guided task is conducted using Mol-Instructions \citep{fang2023mol} (\desctest), an instruction-following dataset of high quality but lacking any quantitative analysis of the designed proteins. For the keyword-guided task, we introduce a new test set, \keywtest, with a datetime cutoff to prevent data contamination. Furthermore, an in-depth analysis is conducted to explore the correlations among various evaluation metrics used in \ourbench.

In summary, the contributions of \ourbench are as follows:
\begin{itemize}[leftmargin=*, itemsep=4pt, parsep=0pt, topsep=0pt]
    \item We present \ourbench, the first comprehensive benchmark for function-guided \denovo protein design, encompassing both description-guided and keyword-guided settings (with 3 fine-grained tasks). We benchmark 8 state-of-the-art models across \mnum metrics spanning 6 dimensions (i.e., Plausibility, Foldability, Language Alignment, Similarity, Novelty, and Diversity).
    \item We analyze and ensure fairness of evaluations. For the description-guided task, we repurpose Mol-Instructions in our benchmark (\desctest). We carefully analyze the testing configuration (by partitioning datasets based on potential overlaps) and demonstrate that fairness of \desctest results are reliable. For keyword-guided tasks, we introduce \keywtest
    which applies a strict datetime cutoff (e.g., only including SwissProt annotations after 2025) to ensure fairness.
    \item We identify key correlations between different metrics that facilitate fairer comparisons and provide insights for future model development. For instance, low PPL and Repeat scores consistently indicate well-folded protein structures (e.g., PPL exhibits Pearson correlations of 0.76 with pLDDT and -0.87 with PAE). Moreover, retrieval-based evaluations are highly sensitive to the chosen retrieval strategy, and although random sampling can be beneficial, absolute values should be interpreted with caution (e.g., the gap in Retrieval Accuracy for natural proteins across different retrieval strategies can reach 66.31\%.).
\end{itemize}

%% file: docs/030Bench.tex
% region tasks and datasets
\subsection{Tasks and Datasets}
\label{sec:task-dataset}

\begin{figure*}[!htb]
    \centering
    \includegraphics[width=1.0\textwidth]{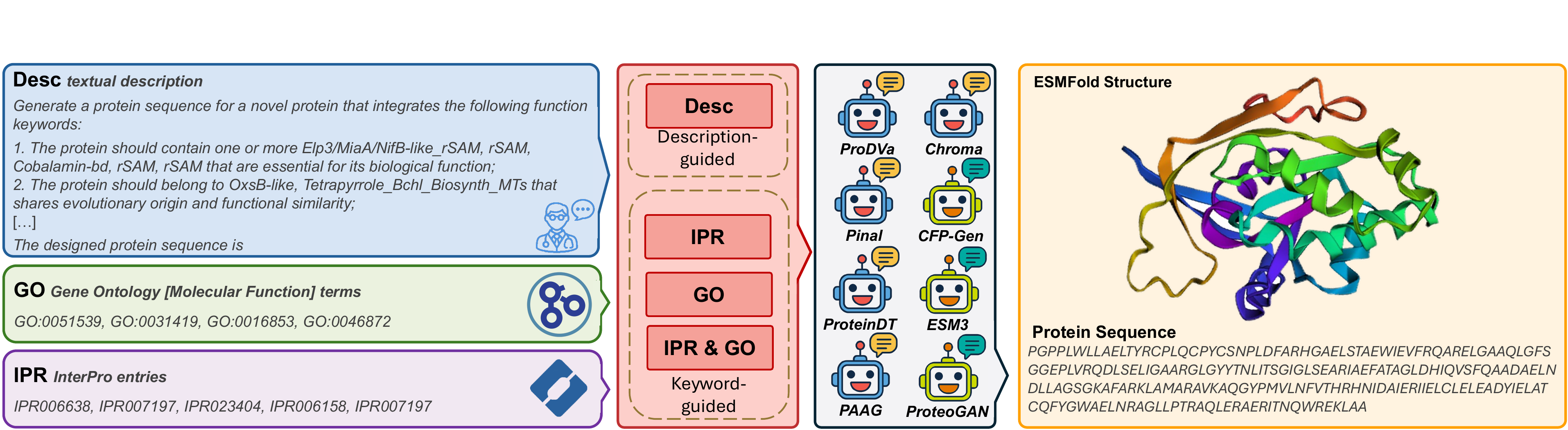}
    \caption{Examples of inputs and outputs for the description-guided protein design task and the keyword-guided protein design task (using GO and/or IPR keywords as inputs). Note that the GO and IPR terms can be converted into textual descriptions. A detailed explanation of this paradigm is provided in Appendix~\ref{sec:app-datasets-keyword}.}
    \label{fig:tasks-example}
\end{figure*}

As demonstrated in Figure \ref{fig:tasks-example}, \denovo protein design from function can be categorized into two types, description-guided and keyword-guided. The objective of both tasks is to generate novel proteins with specific functions, while the input format differs.

\subsubsection{Description-guided Protein Design}

\paragraph{Task Definition}
The description-guided task is to design novel protein \(P\) with function description \(t\) written in natural language. The function description is to describe the overall function of \(P\). The objective of this task is to generate a novel protein using the 20 standard amino acids \(A=\{a_1,a_2,\cdots,a_{20}\}\).
\begin{equation*}
    p(P\mid t) = p((x_1, x_2, \cdots, x_k)\mid t, \forall i, x_i \in A)
\end{equation*}

\paragraph{Dataset}
Mol-Instructions~\citep{fang2023mol} is a diverse, high-quality, large-scale instruction dataset for the biomolecular domain. We use its protein-oriented instruction test set as the evaluation set for the description-guided task, referred to as \desctest. The detailed construction process of \desctest is provided in Appendix~\ref{sec:app-datasets-description}.

\subsubsection{Keyword-guided Protein Design}

\paragraph{Task Definition}
The keyword-guided task is to design a novel protein $P$ based on a set of keywords $K = \{k_1, k_2, \cdots, k_n\}$, where each keyword $k_i$ corresponds to either a Gene Ontology (molecular function) term or an InterPro term in \ourbench. Additionally, the keyword names are converted into textual function descriptions for comparison with models that accept text as input. The objective is to generate a novel protein based on $K$.
\begin{equation*}
    p(P\mid K) = p((x_1, x_2, \cdots, x_k)\mid K, \forall i, x_i \in A)
\end{equation*}

\paragraph{Dataset}
In contrast to the description-guided task, the input for the keyword-guided task consists of a series of keywords. In \ourbench, we manually curated a novel test dataset from UniProt/SwissProt~\cite{uniprot2018uniprot}, termed \keywtest. To prevent potential data leakage, the inclusion period for \keywtest is restricted to January 1, 2025, through August 25, 2025. The detailed construction of \keywtest is presented in Appendix \ref{sec:app-datasets-keyword}.

\subsubsection{Analysis for Fairness Evaluation in Benchmarking}
It is important to ensure that the benchmark evaluation remains fair under specific controlled settings.
To assess the impact of potential data contamination (i.e., overlaps
\footnote{We define hard overlaps as instances where an identical function–sequence pair appears in both the training and test sets. Additionally, soft overlaps refer to instances where the functions or sequences are similar.}
in protein functions or sequences within the training sets of the baseline models), we provide a detailed quantitative analysis in Appendix~\ref{sec:app-datasets-fairness}. We ensure that \ourbench contains no hard overlaps by carefully selecting the test set (the Mol-Instructions test set is excluded from all baseline models) or by curating it (in \keywtest, we apply a datetime cutoff). For soft overlaps, our primary concern is that similar functions may potentially overlap.
In Appendix~\ref{sec:app-datasets-fairness}, we show that only 0.24\% of the proteins in the Mol-Instructions test set share similar functions with at least 50\% functional identity in SwissProt, which is primarily used by the baselines as the training set or its subsets. Further analysis demonstrates that excluding these soft overlaps in similar functions has minimal, or even positive, impact on the experimental results. This indicates that soft overlaps do not compromise the fairness of our evaluation.
% endregion tasks and datasets

% region Baselines
\subsection{Baselines}
\label{sec:baselines}

We ensemble recent \denovo protein design from function baselines as shown in Table \ref{tab:baselines}. For description-guided task, ProDVa~\citep{liu2025proteindesigndynamicprotein}, Pinal~\citep{dai2024toward}, PAAG~\citep{yuan2024annotation}, Chroma~\citep{ingraham2023illuminating} and ProteinDT~\citep{liu2025text} are supported. For keyword-guided task, ESM3~\citep{hayes2025simulating}, CFP-Gen~\citep{yin2025cfpgen} and ProteoGAN~\citep{10.1093/bioinformatics/btac353} are supported. More details about these baseline are listed in Appendix~\ref{sec:app-baselines}.

\input{tabs/main/bench/baseline_desc.tex}

% endregion Baselines

% region Metrics
\subsection{Metrics}
\label{sec:metric}

% region Subsection: Plausibility
\paragraph{Plausibility}
To evaluate the plausibility of the designed proteins, we introduce six metrics.
Direct experimental validation of sequence plausibility is impractical for large-scale studies. Instead, we employ three protein language models—ProtGPT2~\citep{ferruz2022protgpt2}, ProGen2~\citep{nijkamp2023progen2}, and RITA~\citep{hesslow2022rita}—to compute sequence perplexity. These yield \textbf{PPL-ProtGPT2}, \textbf{PPL-ProGen2}, and \textbf{PPL-RITA}, which provide model-based estimates of how well the designed proteins conform to the distributional properties of natural proteins.
Moreover, natural proteins rarely contain long repetitive fragments, yet prior studies~\citep{ferruz2022protgpt2,wang2024diffusion} indicate that generative models may produce sequences with abnormally high repetition, potentially impairing protein functionality. To assess this, we compute \textbf{Rep-2} and \textbf{Rep-5} following Rep-N~\citep{welleck2019neuraltextgenerationunlikelihood}. In addition, we propose a new metric, \textbf{Repeat}, which evaluates the fraction of repetitive sequence fragments from a biologically informed perspective.
% endregion Subsection: Sequence-based

% region Subsection: Foldability
\paragraph{Foldability}
We use ESMFold~\citep{lin2022language} to predict 3D structure for the designed sequence and present two metrics for accessing its foldability. Reasonable proteins must exhibit sufficient foldability to perform their functions. We compute the average predicted local distance difference test (\textbf{pLDDT}) and predicted aligned error (\textbf{PAE}) across the entire structure~\citep{jumper2021highly,akdel2022structural,varadi2022alphafold}.

% endregion Subsection: Foldability

% region Subsection: Language Alignment
\paragraph{Language Alignment}
Function-guided proteins should faithfully reflect user-specified properties. To evaluate the degree of alignment between textual descriptions and designed proteins, we employ 5 metrics: 4 model-based metrics, and 1 retrieval-based metric.
For model-based evaluation, we use ProTrek~\citep{su2024protrek}, a multimodal protein language model pre-trained on large-scale protein–function pairs, to compute the \textbf{ProTrek Score}, defined as the cosine similarity between protein embeddings and function embeddings. In addition, EvoLlama~\citep{liu2024evollama}, fine-tuned on our dataset, is employed to derive the \textbf{EvoLlama Score}, which measures the cosine similarity between the ground-truth function and the function predicted by EvoLlama. 
We calculate keyword-level recovery rates of designed proteins relative to natural proteins, specifically \textbf{IPR Recovery} and \textbf{GO Recovery}. Following the setting of CFP-Gen and ESM3, The IPR annotations are obtained using InterProScan~\citep{paysan2023interpro}, while GO terms are derived from DeepGO-SE~\citep{kulmanovProteinFunctionPrediction2024}. Higher recovery rates indicate stronger preservation of functional characteristics.
For retrieval-based evaluation, we follow previous work~\citep{liu2025text} to compare each designed sequence against its ground-truth function and \( T - 1\) randomly selected functions using ProTrek. \textbf{Retrieval Accuracy} is defined as the proportion of cases in which the true function–sequence pair ranks highest among all comparisons.
% endregion Language Alignment

% region Subsection: Similarity
\paragraph{Similarity}
Proteins with analogous sequences or structures often perform similar functions.
\textbf{ESMScore} is a composite metric comprising ESM-F1, ESM-Precision, and ESM-Recall, which measures sequence similarity between designed sequences and ground truth using ESM-2-650M~\citep{lin2022language}, following the formulation of BERTScore~\citep{bert-score}.
\textbf{GT-Identity} denotes the sequence similarity between a designed sequence and the ground truth, computed with MMseqs2~\citep{kallenborn2024gpu}, which employs multiple sequence alignment (MSA) to quantify similarity.
Additionally, we assess structural similarity between designed proteins and ground truth using TMscore~\citep{zhang2004scoring}, denoted as \textbf{GT-TMScore}.
% endregion Similarity

% region Subsection: Novelty
\paragraph{Novelty}
\textbf{Novelty} represents the dissimilarity of designed proteins with respect to reference databases. Here, we assess novelty from both sequence and structure perspectives. We employ MMseqs2 to calculate the novelty of designed sequences within UniProtKB~\citep{uniprot2018uniprot}, denoted as \textbf{Novelty-Seq}. Additionally, we utilize Foldseek~\citep{kimRapidSensitiveProtein2025} to evaluate the novelty of designed protein structures within AlphaFoldDB/SwissProt, denoted as \textbf{Novelty-Struct}.
% endregion Novelty

% region Subsection: Diversity
\paragraph{Diversity}
Diversity evaluates whether the model generates diverse proteins rather than producing minor variations of a template sequence, by computing the average pairwise dissimilarity among all proteins designed for the same function. Similar to Novelty, we use MMseqs2 to compute sequence diversity (\textbf{Diversity-Seq}) and Foldseek to compute structural diversity (\textbf{Diversity-Struct}).
% endregion Diversity

% endregion Metrics 

%% file: tabs/main/bench/baseline_desc.tex
\begin{table*}[!htb]
    \caption{Comparison with various baselines. The symbols \ \cmark\  and \ \xmark\  denote "Supported" and "Not Supported," respectively. As ProteoGAN, ESM3, and CFP-Gen offer only partial support for IPR and GO keywords, we report both the number of keywords each model claims to support and the number of keywords available in the \keywtest .}
    \label{tab:baselines}
    \centering
    \resizebox{\textwidth}{!}{
    \begin{tabular}{lcccccc p{1.0\textwidth}}
    \toprule \multirow{2}{*}[-0.5ex]{Baselines} 
    & \multicolumn{3}{c}{Keyword} 
    & \multirow{2}{*}[-0.5ex]{Desc} 
    & \multicolumn{2}{c}{Training Set} 
    & \multirow{2}{*}[-0.5ex]{Brief Summaries} \\
    \cmidrule[0.5pt](lr){2-4}
    \cmidrule[0.5pt](lr){6-7}
    & {\small GO} & {\small IPR} & {\small GO\&IPR} & & {\small Size} & {\small Access} & \\
    \midrule 
    % ESM3
    \multirow{2}{*}{ESM3} &
    \multirow{2}{*}{\large{\xmark}} &
    \multirow{2}{*}{\makecell{\large{\cmark} \\ 29026\,(1197)}} &
    \multirow{2}{*}{\large{\xmark}} &
    \multirow{2}{*}{\large{\xmark}} &
    \multirow{2}{*}{455M} &
    \multirow{2}{*}{\large{\xmark}} &
    A frontier multimodal generative model tokenizing sequence, structure, and function in unified space, trained with masked language modeling across modalities. \\[3ex]
    % ProteoGAN
    \multirow{2}{*}{ProteoGAN} &
    \multirow{2}{*}{\makecell{\large{\cmark} \\ 50\,(41)}} &
    \multirow{2}{*}{\large{\xmark}} &
    \multirow{2}{*}{\large{\xmark}} &
    \multirow{2}{*}{\large{\xmark}} &
    \multirow{2}{*}{158K} &
    \multirow{2}{*}{\large{\xmark}} &
    Conditional GAN guided by Gene Ontology labels, designed for general-purpose protein generation and evaluated with biologically/statistically inspired metrics. \\[3ex]
    % CFP-Gen
    \multirow{2}{*}{CFP-Gen} &
    \multirow{2}{*}{\makecell{\large{\cmark} \\ 375\,(92)}} &
    \multirow{2}{*}{\makecell{\large{\cmark} \\ 1154\,(170)}} &
    \multirow{2}{*}{\makecell{\large{\cmark}}} &
    \multirow{2}{*}{\large{\xmark}} &
    \multirow{2}{*}{244K} &
    \multirow{2}{*}{\large{\cmark}} &
    Diffusion-based multimodal generator introducing AGFM and RCFE modules, enabling precise functional, structural, and residue-level control for multifunctional protein design. \\
    \midrule 
    % ProteinDT
    \multirow{2}{*}{ProteinDT} &
    \multirow{2}{*}{\large{\cmark}} &
    \multirow{2}{*}{\large{\cmark}} &
    \multirow{2}{*}{\large{\cmark}} &
    \multirow{2}{*}{\large{\cmark}} &
    \multirow{2}{*}{541K} & 
    \multirow{2}{*}{\large{\cmark}} &
    Multimodal framework using ProteinCLAP for joint embeddings, mapping text to protein representations, followed by autoregressive decoding for sequence generation. \\[3ex]
    % Chroma
    \multirow{2}{*}{Chroma} &
    \multirow{2}{*}{\large{\cmark}} &
    \multirow{2}{*}{\large{\cmark}} &
    \multirow{2}{*}{\large{\cmark}} &
    \multirow{2}{*}{\large{\cmark}} &
    \multirow{2}{*}{45K} &
    \multirow{2}{*}{\large{\xmark}} &
    Diffusion-based protein designer integrating polymer physics, enabling programmable design through composable conditioners to enforce multiple constraints. \\[3ex]
    % PAAG
    \multirow{2}{*}{PAAG} &
    \multirow{2}{*}{\large{\cmark}} &
    \multirow{2}{*}{\large{\cmark}} &
    \multirow{2}{*}{\large{\cmark}} &
    \multirow{2}{*}{\large{\cmark}} &
    \multirow{2}{*}{129K} &
    \multirow{2}{*}{\large{\cmark}} &
    A multimodal design model that employs a multilevel alignment module to align sequences and descriptions at both global and local levels. \\[3ex]
    % Pinal
    \multirow{2}{*}{Pinal} &
    \multirow{2}{*}{\large{\cmark}} &
    \multirow{2}{*}{\large{\cmark}} &
    \multirow{2}{*}{\large{\cmark}} &
    \multirow{2}{*}{\large{\cmark}} &
    \multirow{2}{*}{1.7B} &
    \multirow{2}{*}{\large{\xmark}} &
    A large-scale framework, pretrained on 1.7B function–sequence pairs, most of which are synthetic, designing sequence mediated by structure.\\[3ex]
    % ProDva
    \multirow{2}{*}{ProDVa} &
    \multirow{2}{*}{\large{\cmark}} &
    \multirow{2}{*}{\large{\cmark}} &
    \multirow{2}{*}{\large{\cmark}} &
    \multirow{2}{*}{\large{\cmark}} &
    \multirow{2}{*}{640K} &
    \multirow{2}{*}{\large{\cmark}} &
    A multimodal framework integrating a text encoder, a protein language model, and a fragment encoder that dynamically retrieves relevant fragments based on the specified function. \\
    \bottomrule
    \end{tabular}
    }
\end{table*}

%% file: docs/050Results.tex
In this section, we conduct a comprehensive evaluation of these baselines across the two tasks (four fine-grained tasks). The detailed experimental settings are shown in Section \ref{sec:app-res-set}.

\input{tabs/main/results/results_desc.tex}

% region Description-guided
\subsection{Description-Guided}
In Table \ref{tab:main-res-desc}, we report the benchmark results for the description-guided task on 11 main metrics. The complete results are displayed in Appendix \ref{sec:app-res-desc}. Findings are summarized as follows:

\noindent \textbf{(1) ProDVa can design relatively plausible sequence.} Good sequence plausibility is fundamental to foldability and language alignment. The sequences designed by ProDVa exhibit repeat scores exceeding those of natural proteins and suboptimal perplexity scores in sequence rationality, indicating that ProDVa's design modules are capable of generating reasonable protein sequences.

\noindent \textbf{(2) ProDVa and Pinal Generate Foldable Proteins.} Both ProDVa and Pinal achieve substantially higher foldability scores compared with all other models. Specifically, ProDVa reaches the best pLDDT (76.86) and lowest PAE (8.66), while Pinal follows closely with pLDDT (75.25) and PAE (10.96). These results suggest that the sequences produced by both models are structurally stable and more likely to fold into valid conformations, highlighting the effectiveness of their design modules in capturing the structural constraints of proteins.

\noindent \textbf{(3) ProDVa and Pinal exhibit comparable performance in Language Alignment, whereas the remaining baselines demonstrate substantially inferior results.} For language alignment, both models outperform all baselines by large margins. ProDVa achieves the best retrieval accuracy (66.83), while Pinal attains a comparable score (63.43). They also perform significantly better in ProTrek Score and EvoLlama Score compared to ProteinDT, Chroma, and PAAG. This indicates that the semantic and evolutionary information embedded in the descriptions are effectively translated into protein sequences by ProDVa and Pinal, whereas the baseline methods fail to capture such alignment.

\noindent \textbf{(4) ProDVa and Pinal perform poorly with respect to novelty and diversity.} While excelling in plausibility, foldability, and alignment, both models show relatively low novelty and diversity compared with baselines. The low novelty scores (ProDVa Seq/Struct: 14.64/36.31, Pinal Seq/Struct: 43.82/17.23) suggest that the designed sequences tend to remain close to the natural protein landscape. At the same time, their low diversity scores (ProDVa Seq/Struct: 46.10/52.592; Pinal Seq/Struct: 75.88/72.73) indicate that the models may confine functional design to narrow clusters in sequence/structure space. This reflects a trade-off: in order to achieve better functional alignment, ProDVa and Pinal may sacrifice exploration of diverse solutions, thereby limiting their coverage of the broader protein landscape.

% endregion Description-guided

% region Keyword-guided
\input{tabs/main/results/results_keyw.tex}

\subsection{Keyword-Guided}
As illustrated in Table \ref{tab:main-res-keyw}, we report the benchmark results for keyword-guided task on 12 main metrics. The complete results of keyword-guided task on all metrics are in Appendix \ref{sec:app-res-keyw}. Based on these results, our key findings are as follows:

\noindent \textbf{(1) CFP-Gen, Pinal and ESM3 show great performance in Perplexity while the Repeat show poorly.} 
These models achieve the lowest perplexity scores (CFP-Gen: 4.94–5.23; Pinal: 6.85–8.12; ESM3: 6.33), indicating that their generated sequences exhibit strong rationality under the protein language model. However, they also show much higher Repeat (ranging from 11.86 to 28.13) compared with ProDVa or Chroma, suggesting that the improved plausibility comes at the cost of local redundancy in sequence design.

\noindent \textbf{(2) CFP-Gen, ProDVa and Pinal can design foldable proteins.} 
These models consistently achieve high pLDDT and low PAE across different evaluation settings. ProDVa stands out with the best overall foldability (pLDDT: 72.80–74.73; PAE: 6.11–8.06), while CFP-Gen and Pinal also produce structures with good confidence (pLDDT around 69–76; PAE around 11–14). This indicates that their design strategies are particularly effective at generating sequences that fold into stable 3D structures.

\noindent \textbf{(3) CFP-Gen shows great performance among the keyword-guided baselines, while weak performance among the description-guided baselines.} 
Compared to other keyword-guided baselines such as ProteinDT, Chroma, or PAAG, CFP-Gen achieves significantly higher alignment with biological annotations (e.g., IPR Recovery up to 35.21 and GO Recovery up to 21.05). In contrast, its performance was still less competitive than Pinal and ProDVa.

\noindent \textbf{(4) Baselines perform better on IPR-guided task than GO-guided task.} 
In the single-keyword setting, models achieve higher recovery performance on the metric that matches the input type. Moreover, using IPR as input generally leads to stronger performance across Plausibility, Foldability, and most Language Alignment metrics, indicating that IPR annotations provide more precise constraints for protein design than GO terms.  

\noindent \textbf{(5) The IPR\&GO-guided task imposes stricter constraints than the other two tasks, while less than description-guided task.} 
When extending from single- to dual-keyword guidance, we observe a nuanced trade-off. IPR Recovery increases while GO Recovery decreases, suggesting that IPR contributes more strongly to functional alignment in this joint setting. Meanwhile, both ProTrek Score and Retrieval Accuracy are improved, while Plausibility and Foldability remain largely unchanged. These results imply that combining IPR and GO constraints reduces the design difficulty in terms of language alignment, enabling models to better converge on functionally consistent sequences without sacrificing structural quality. Nevertheless, the alignment scores are still higher than those obtained in the description-guided setting, showing that structured keyword guidance provides clearer signals for functional targeting, albeit at the cost of reduced novelty and diversity. 

% endregion Keyword-guided 

%% file: tabs/main/results/results_desc.tex
\begin{table*}[!htb]
    \caption{Benchmark results for the description-guided task (\firsttext{Best}, \secondtext{Second Best}, \thirdtext{Third Best}). - indicates not applicable.}
    \label{tab:main-res-desc}
    \centering
    \resizebox{\textwidth}{!}{
    \begin{tabular}{lccccccccccc}
        \toprule
        \multirow{3}{*}{\textbf{Models}} &
        \multicolumn{2}{c}{\textbf{Plausibility}} & 
        \multicolumn{2}{c}{\textbf{Foldability}} & 
        \multicolumn{3}{c}{\textbf{Language Alignment \% } } &
        \multicolumn{2}{c}{\textbf{Novelty \% }} &
        \multicolumn{2}{c}{\textbf{Diversity \% }} \\
        \cmidrule(lr){2-3} \cmidrule(lr){4-5}\cmidrule(lr){6-8}\cmidrule(lr){9-10}\cmidrule(lr){11-12}
        & Perplexity $\downarrow$ &
        Repeat \% $\downarrow$ &
        pLDDT $\uparrow$ & 
        PAE $\downarrow$ & 
        \makecell{ProTrek\\Score $\uparrow$} &
        \makecell{EvoLlama\\Score $\uparrow$} &
        \makecell{Retrieval\\Accuracy $\uparrow$} & 
        Seq $\uparrow$ & 
        Struct $\uparrow$ &
        Seq $\uparrow$ & 
        Struct $\uparrow$ \\
        \midrule
        \arrayrulecolor{gray!20}
        Natural & 5.99 & 1.99 & 80.64 & 9.20 & 27.00 & 60.33 & 89.01 & 4.90 & 13.56 & - & - \\
        \midrule
        Random(U) & 21.71 & 0.72 & 22.96 & 24.85 & 1.03 & 36.22 & 13.03 & 58.14 & 77.64 & 98.61 & 81.59 \\
        Random(E) & 18.68 & 1.15 & 25.77 & 24.71 & 1.04 & 34.11 & 12.73 & 60.19 & 76.82 & 99.74 & 81.45 \\
        \midrule
        ProteinDT & 12.41 & 6.83 & 38.29 & 25.13 & 1.20 & \third{40.57} & \third{16.92} & \first{70.74} & \second{71.16} & \third{99.31} & \first{83.67} \\
        Chroma & \third{12.19} & \third{2.59} & \third{59.18} & \third{15.03} & \third{2.10} & 40.10 & 13.01 & \third{58.68} & \third{51.06} & \first{98.42} & \third{79.90} \\
        PAAG & 17.84 & \second{2.34} & 28.39 & 25.38 & 1.29 & 34.39 & 13.43 & \second{63.64} & \first{77.34} & \second{99.41} & \second{82.16} \\
        Pinal & \first{5.81} & 12.83 & \second{75.25} & \second{10.96} & \first{17.50} & \first{53.40} & \second{63.43} & 43.82 & 17.23 & 75.88 & 72.73 \\
        ProDVa & \second{7.63} & \first{1.92} & \first{76.86} & \first{8.66} & \second{17.40} & \second{51.19} & \first{66.83} & 14.64 & 36.31 & 46.10 & 52.59 \\
        \arrayrulecolor{black}
        \bottomrule
    \end{tabular}
    }
\end{table*}

%% file: tabs/main/results/results_keyw.tex
\begin{table*}[!htb]
    \caption{Benchmark results for the keyword-guided task (\firsttext{Best}, \secondtext{Second Best}, \thirdtext{Third Best}). - indicates not applicable.}
    \label{tab:main-res-keyw}
    \centering
    \resizebox{\textwidth}{!}{
    \begin{tabular}{lcccccccccccc}
        \toprule
        \multirow{3}{*}{\textbf{Models}} &
        \multicolumn{2}{c}{\textbf{Plausibility}} & 
        \multicolumn{2}{c}{\textbf{Foldability}} & 
        \multicolumn{4}{c}{\textbf{Language Alignment \% } } &
        \multicolumn{2}{c}{\textbf{Novelty \% }} &
        \multicolumn{2}{c}{\textbf{Diversity \% }} \\
        \cmidrule(lr){2-3} \cmidrule(lr){4-5} \cmidrule(lr){6-9}\cmidrule(lr){10-11} \cmidrule(lr){12-13}
        & Perplexity $\downarrow$ &
        Repeat \% $\downarrow$ &
        pLDDT $\uparrow$ & 
        PAE $\downarrow$ & 
        \makecell{ProTrek\\Score $\uparrow$} &
        IPR Recovery$\uparrow$ & 
        GO Recovery$\uparrow$ &
        \makecell{Retrieval\\Accuracy $\uparrow$} & 
        Seq $\uparrow$ & 
        Struct $\uparrow$ &
        Seq $\uparrow$ & 
        Struct $\uparrow$ \\
        \midrule
        \multicolumn{13}{c}{\textbf{\textit{guided with GO keywords}}} \\
        \midrule
        \arrayrulecolor{gray!20}
        Natural & 9.17 & 2.17 & 76.92 & 10.54 & 21.60 & 100.00 & 100.00 & 77.49 & 4.07 & 18.15 & - & - \\
        \midrule
        Random(U) & 21.74 & 0.72 & 23.20 & 24.56 & 4.29 & 0.00 & 20.79 & 10.00 & 58.04 & 76.75 & 97.75 & 81.56 \\
        Random(E) & 18.68 & 1.14 & 25.99 & 24.47 & 3.44 & 0.00 & 11.71 & 11.06 & 60.28 & 75.92 & 99.35 & 81.54 \\
        \midrule
        ProteoGAN & 18.03 & \third{2.50} & 28.72 & 24.67 & 4.42 & 0.00 & 14.99 & 13.84 & \second{65.24} & \first{75.82} & \second{99.31} & \second{84.37} \\
        CFP-Gen & \first{5.16} & 12.67 & \second{73.38} & \third{14.61} & \third{10.03} & \third{9.67} & 18.98 & \third{38.87} & 47.85 & 28.28 & 69.84 & \third{81.76} \\
        ProteinDT & 12.23 & 7.98 & 40.35 & 25.57 & 1.70 & 0.03 & 18.52 & 15.39 & \first{75.41} & \second{74.62} & \first{99.79} & \first{84.53} \\
        Chroma & 12.18 & 2.71 & 59.27 & 15.00 & 1.84 & 0.23 & 16.33 & 12.07 & 59.35 & 50.88 & 97.61 & 79.79 \\
        PAAG & 18.08 & \second{2.48} & 31.47 & 23.88 & 4.38 & 0.00 & \third{21.66} & 16.45 & \third{62.36} & \third{73.36} & \third{98.33} & 81.73 \\
        Pinal & \second{6.85} & 14.13 & \third{72.58} & \second{11.79} & \second{12.69} & \second{19.26} & \second{22.76} & \second{49.93} & 46.06 & 19.27 & 88.75 & 79.00 \\
        ProDVa & \third{11.16} & \first{1.87} & \first{74.73} & \first{6.11} & \first{14.42} & \first{20.22} & \first{30.24} & \first{52.38} & 25.02 & 32.72 & 42.84 & 32.02 \\
        \bottomrule
        \arrayrulecolor{black}
        \midrule
        \multicolumn{13}{c}{\textbf{\textit{guided with IPR keywords}}} \\
        \midrule
        \arrayrulecolor{gray!20}
        Natural & 9.73 & 2.23 & 75.77 & 11.13 & 25.29 & 83.22 & 100.00 & 83.22 & 4.47 & 20.09 & - & - \\
        \midrule
        Random(U) & 21.76 & 0.69 & 23.40 & 24.42 & 7.53 & 0.00 & 25.75 & 11.72 & 57.21 & 76.47 & 97.82 & 81.56 \\
        Random(E) & 18.67 & 1.16 & 26.29 & 24.34 & 6.11 & 0.00 & 13.06 & 12.80 & 59.57 & 75.05 & 99.35 & 81.46 \\
        \midrule
        ESM3 & \second{6.33} & 28.13 & 60.90 & 16.73 & 6.22 & 20.17 & 15.43 & 33.01 & \second{71.87} & 37.56 & 93.69 & 76.79 \\
        CFP-Gen & \first{4.94} & 11.86 & \first{76.36} & \second{12.54} & \third{10.21} & \first{32.79} & \second{23.41} & \third{40.96} & 49.46 & 23.15 & 82.09 & \second{82.08} \\
        ProteinDT & 11.87 & 10.02 & 37.59 & 26.19 & 3.85 & 0.08 & \third{20.76} & 16.13 & \first{73.57} & \second{76.03} & \first{99.85} & \first{84.81} \\
        Chroma & 12.17 & \third{2.60} & 59.76 & 14.67 & 3.82 & 0.17 & 17.15 & 13.68 & 59.25 & \third{50.77} & \third{97.73} & 79.88 \\
        PAAG & 17.85 & \second{2.32} & 30.89 & 24.98 & 5.98 & 0.08 & 13.85 & 14.37 & \third{64.71} & \first{79.23} & \second{99.40} & \third{81.48} \\
        Pinal & \third{8.12} & 16.73 & \third{65.69} & \third{14.10} & \second{14.38} & \second{25.63} & 15.93 & \first{57.59} & 51.61 & 27.00 & 89.42 & 80.38 \\
        ProDVa & 12.47 & \first{1.99} & \second{72.80} & \first{6.86} & \first{15.19} & \third{24.58} & \first{26.59} & \second{51.99} & 28.86 & 31.26 & 50.11 & 43.44 \\
        \bottomrule
        \arrayrulecolor{black}
        \midrule
        \multicolumn{13}{c}{\textbf{\textit{guided with IPR\&GO keywords}}} \\
        \midrule
        \arrayrulecolor{gray!20}
        Natural & 8.96 & 2.16 & 77.17 & 10.48 & 27.36 & 87.39 & 100.00 & 87.39 & 3.89 & 17.72 & - & - \\
        \midrule
        Random(U) & 21.72 & 0.73 & 22.85 & 24.72 & 4.84 & 0.00 & 25.38 & 9.45 & 57.48 & 77.18 & 97.75 & 81.73 \\
        Random(E) & 18.68 & 1.14 & 25.60 & 24.59 & 3.72 & 0.00 & 14.67 & 11.28 & 60.06 & 75.98 & 99.29 & 81.57 \\
        \midrule
        CFP-Gen & \first{5.23} & 13.14 & \second{72.70} & \third{14.45} & \third{11.68} & \first{35.21} & \second{23.31} & \third{45.77} & 54.72 & 28.89 & 75.16 & \second{81.91} \\
        ProteinDT & 12.81 & 6.81 & 36.46 & 25.75 & 3.06 & 0.36 & 15.92 & 19.29 & \first{71.44} & \second{75.73} & \first{99.67} & \first{84.18} \\
        Chroma & 12.19 & \second{2.53} & 58.71 & 15.33 & 2.19 & 0.16 & 14.12 & 11.67 & \third{59.36} & \third{51.45} & \third{97.85} & 79.97 \\
        PAAG & 17.80 & \first{2.32} & 30.05 & 25.69 & 4.66 & 0.02 & 9.77 & 11.82 & \second{65.07} & \first{81.53} & \second{99.42} & \third{81.51} \\
        Pinal & \second{7.39} & 16.22 & \third{69.32} & \second{12.97} & \second{15.26} & \second{33.08} & \third{21.64} & \second{60.88} & 49.03 & 22.43 & 85.67 & 78.20 \\
        ProDVa & \third{10.48} & \third{2.61} & \first{74.26} & \first{8.06} & \first{16.78} & \third{30.95} & \first{25.24} & \first{61.23} & 21.97 & 24.20 & 51.36 & 51.35 \\
        \arrayrulecolor{black}
        \bottomrule
    \end{tabular}
    }
\end{table*}

%% file: docs/060Discussion.tex
In this section, we provide a comprehensive analysis regarding the correlations among different evaluation metrics on \ourbench.

\subsection{Does PPL Accurately Reflect pLDDT and PAE?}
\label{sec:discuss-ppl-plddt}

\begin{figure}[!htb]
    \centering

    \begin{subfigure}[b]{0.49\linewidth}
        \centering
        \includegraphics[width=\linewidth]{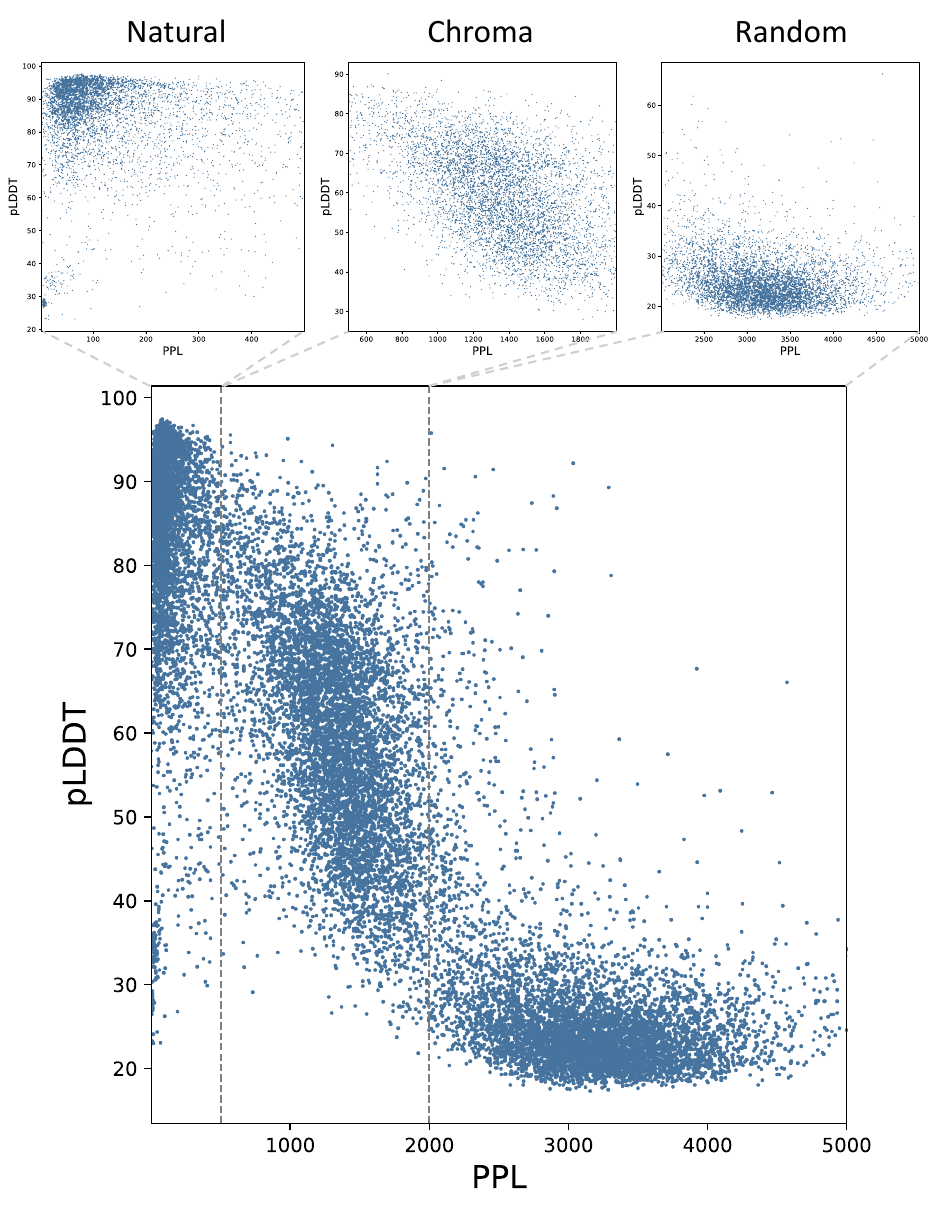}
        \caption{}
    \end{subfigure}
    \hfill
    \begin{subfigure}[b]{0.49\linewidth}
        \centering
        \includegraphics[width=\linewidth]{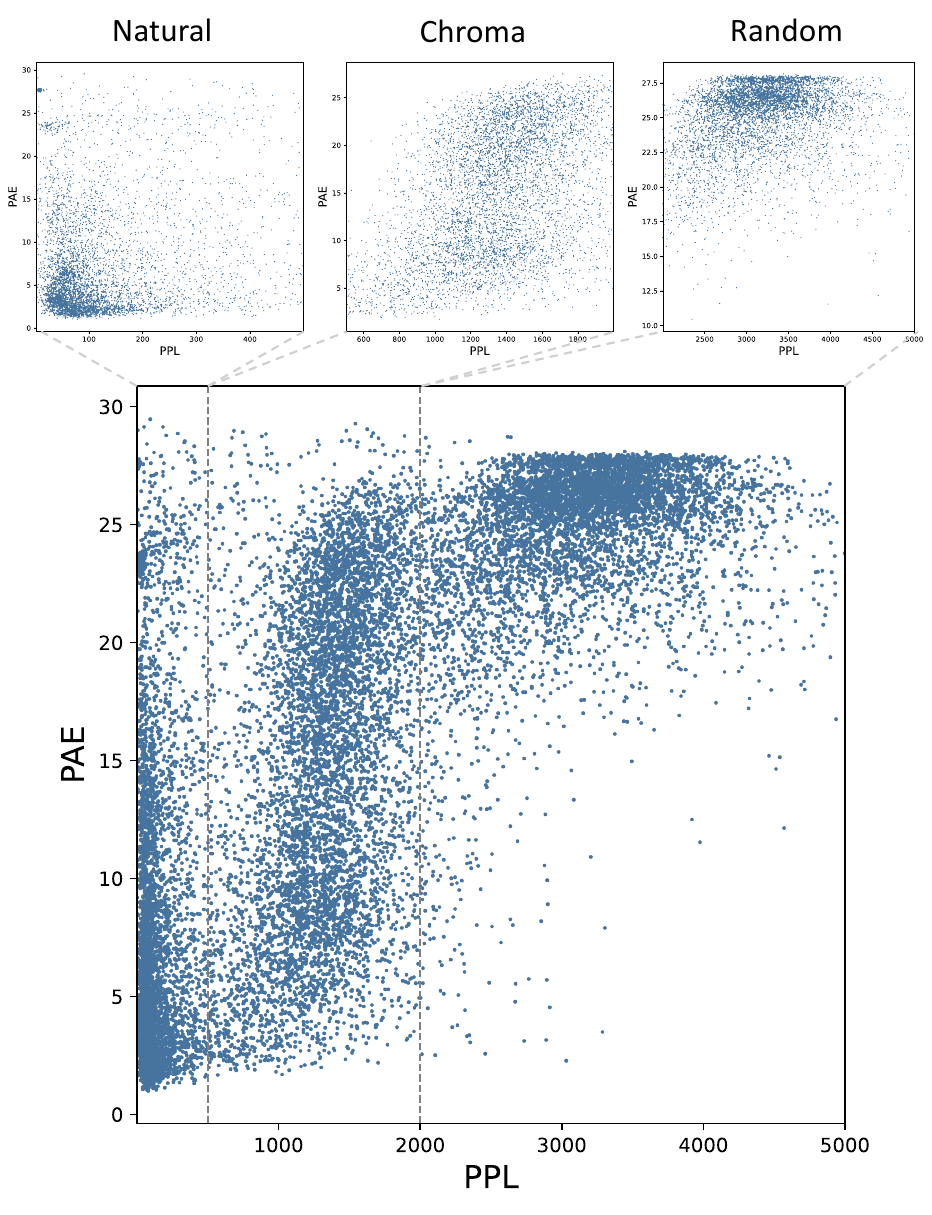}
        \caption{}
    \end{subfigure}

    \vspace{0.3em}

    \begin{subfigure}[b]{0.49\linewidth}
        \centering
        \includegraphics[width=\linewidth]{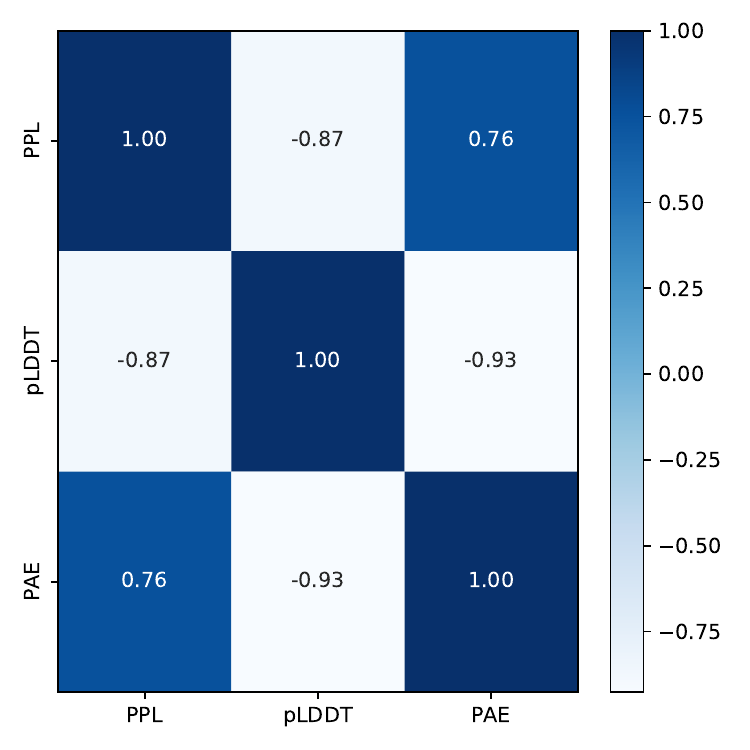}
        \caption{}
    \end{subfigure}

    \caption{\textbf{(a)} presents the distribution of PPL and pLDDT. \textbf{(b)} displays the distribution of PPL and PAE. \textbf{(c)} illustrates the Pearson correlation among these metrics. PPLs are categorized into low, medium, and high ranges: below 500, 500--2,000, and above 2,000, respectively.}
    \label{fig:ppl-plddt-pae}
    % \vspace{-0.8em} 
\end{figure}

We begin by exploring to what extent sequence-level metrics reflect protein structures. Folding proteins into 3D structures using AlphaFold \citep{jumper2021highly} or ESMFold \citep{lin2022language} is time-consuming and requires substantial computational resources, particularly for longer sequences. Previous studies \citep{hesslow2022rita, ferruz2022protgpt2} have observed a correlation between PPL and pLDDT. However, no empirical results or further analyses have been conducted to investigate the correlation.

Results are randomly sampled from natural proteins with low PPL, Chroma-designed proteins with medium PPL, and randomly generated proteins with high PPL. Figure \ref{fig:ppl-plddt-pae} presents the distributions of PPL, pLDDT, and PAE. Proteins with high pLDDT are predominantly clustered in the low PPL range, whereas those with low pLDDT are concentrated in the high PPL range. For the proteins situated between these two clusters, a negative correlation is observed between PPL and pLDDT. Specifically, lower PPL are generally associated with higher pLDDT. A similar pattern is observed in the distribution of PPL and PAE. Therefore, we empirically categorize PPL into three ranges, denoted as low PPL range (below 500), medium PPL range (between 500 and 2,000), and high PPL range (above 2,000). Additionally, the Pearson correlation \citep{cohen2009pearson} in Figure \ref{fig:ppl-plddt-pae}(c) highlights the relationships between PPL, pLDDT, and PAE.

\takeaway{In the low PPL range, proteins are well-folded, exhibiting high pLDDT and low PAE. In contrast, proteins in the high PPL range struggle to fold into plausible structures. Within the medium PPL range, proteins with higher PPL tend to display lower pLDDT and higher PAE.}

\subsection{Do Repetitive Patterns Lead to Lower Structural Plausibility?}

\begin{figure}[!htb]
    \centering
    \begin{subfigure}[b]{0.49\linewidth}
        \centering
        \includegraphics[width=\linewidth]{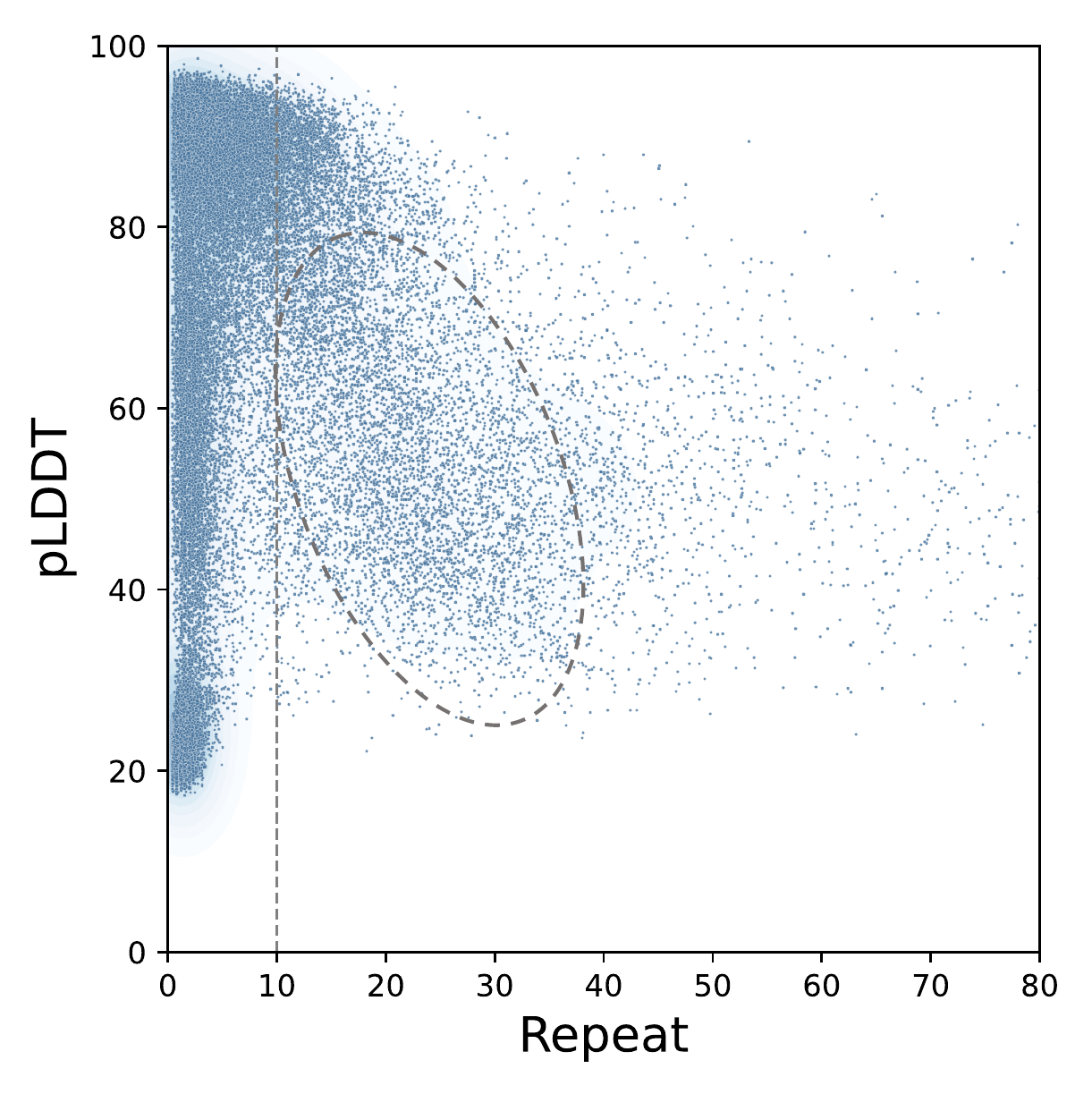}
        \caption{}
    \end{subfigure}
    \hfill
    \begin{subfigure}[b]{0.49\linewidth}
        \centering
        \includegraphics[width=\linewidth]{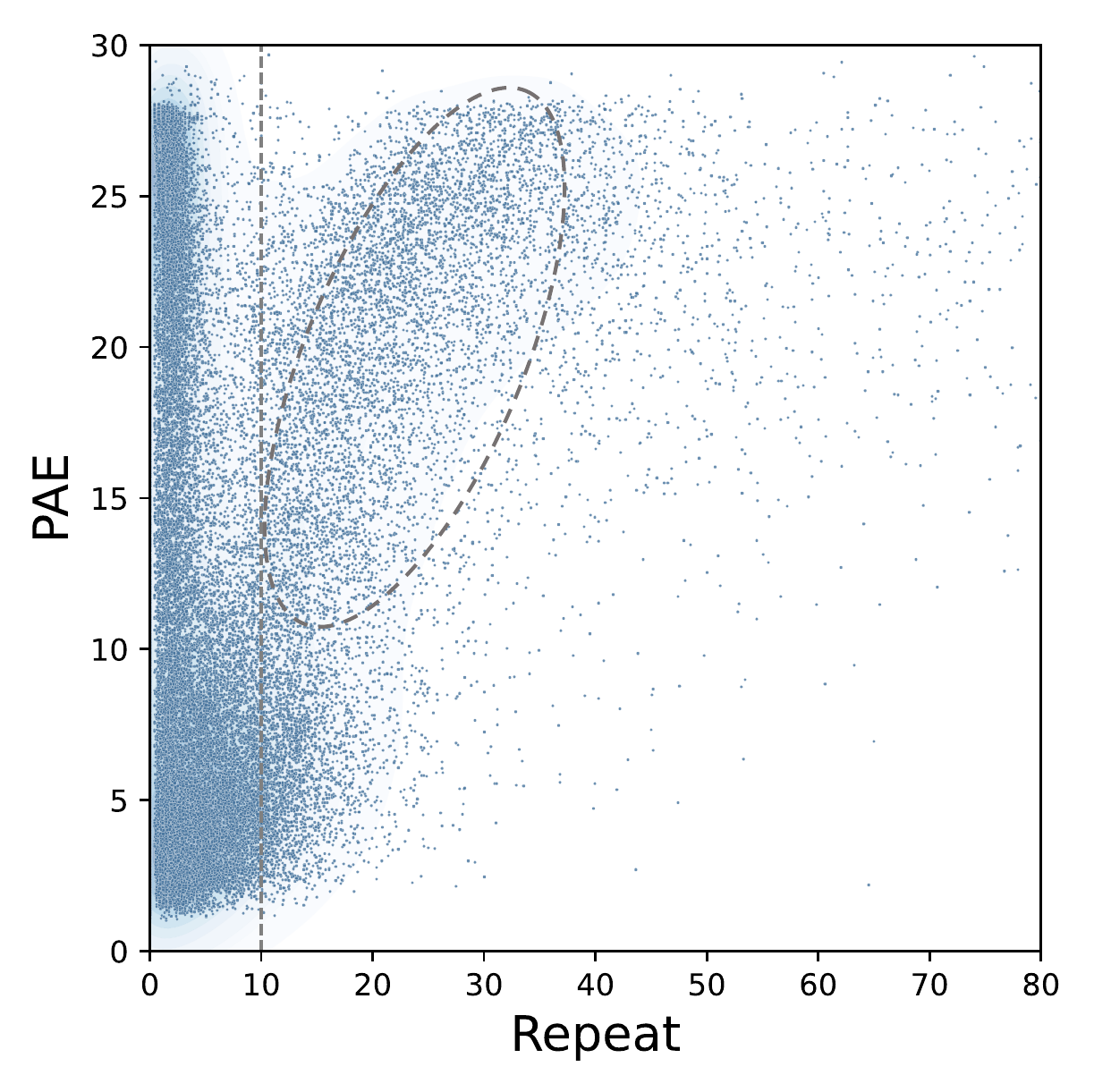}
        \caption{}
    \end{subfigure}
    \caption{\textbf{(a)} Distribution of Repeat and pLDDT. \textbf{(b)} Distribution of Repeat and PAE.}
    \label{fig:rep-plddt-pae}
\end{figure}

Previous research \citep{wang2024diffusion} has found that repetitive patterns occurring in amino acid sequences may result in low pLDDT, thereby leading to lower structural plausibility. We conduct an empirical analysis using the Repeat metric to measure the correlation between this pattern and foldability. Proteins designed by ESM3 and Pinal exhibit repetitive patterns, as indicated by their high scores on the Repeat metric. Figure \ref{fig:rep-plddt-pae} presents the distribution of Repeat and pLDDT and PAE for proteins randomly sampled from natural sequences and those designed by ESM3 and Pinal. One observation is that when the Repeat remains relatively low, there is no clear relationship between Repeat and foldability. In other words, a low Repeat does not necessarily indicate that a protein is well-folded. However, when the Repeat exceeds 10, higher Repeat are associated with lower pLDDT and higher PAE. Therefore, it is important to maintain repetitive patterns below a certain threshold (e.g., Repeat < 10) when designing well-folded proteins.

\takeaway{High Repeat (typically above 10) in protein sequences are associated with lower structural plausibility as indicated by lower pLDDT and higher PAE.}

\subsection{How Faithfully Do Designed Proteins Align with Functional Descriptions?}

\begin{figure}[!htb]
    \centering
    \includegraphics[width=0.5\linewidth]{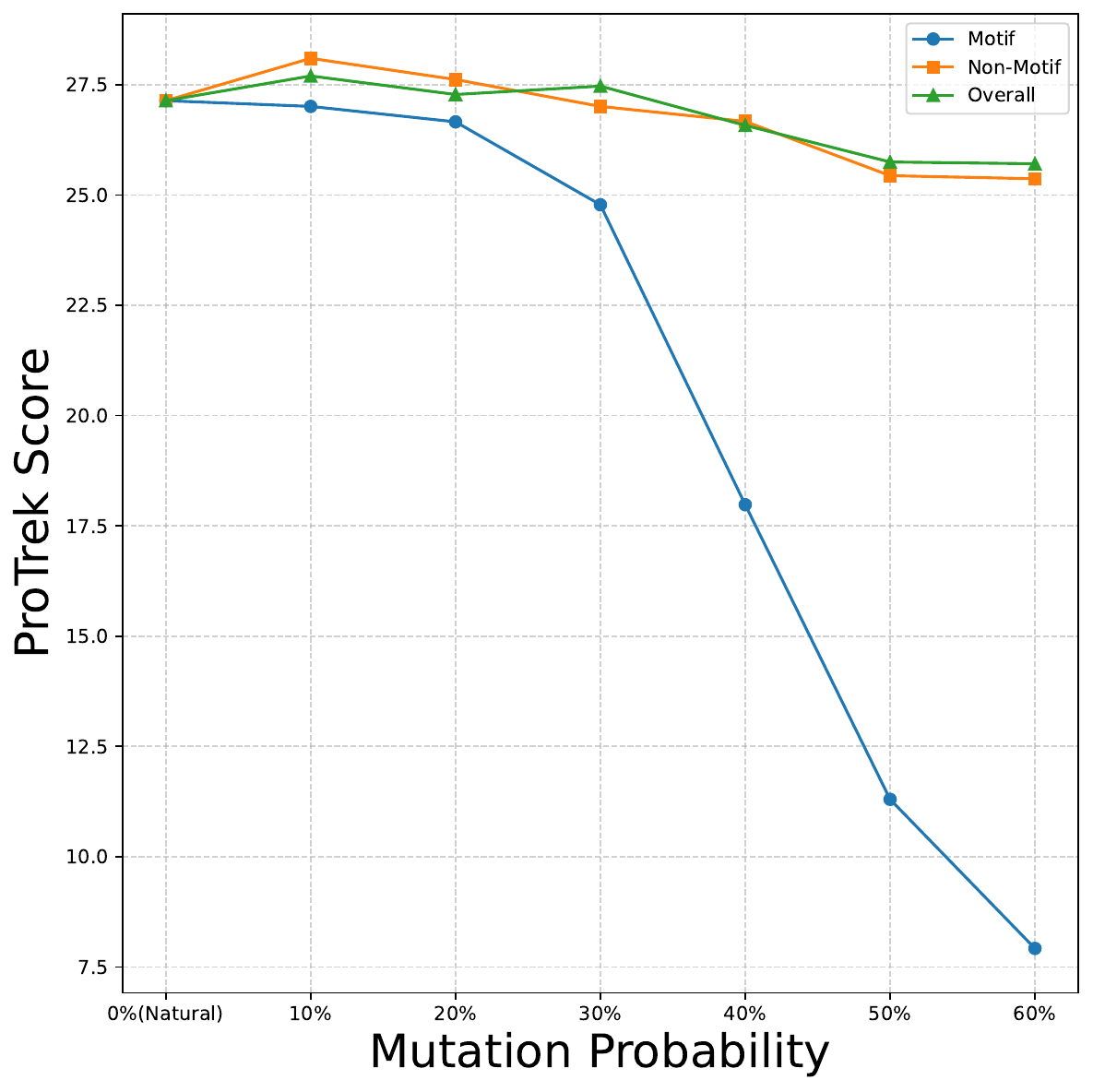}
    \caption{Results for random mutations in natural proteins. Motif and Non-motif indicate mutations within or outside motif regions. Overall includes all mutations.}
    \label{fig:motif-sen}
\end{figure}

The most reliable strategy for evaluating the alignment between designed proteins and input textual descriptions is through wet-lab experiments. However, such experiments are time-consuming and costly. Therefore, employing computational methods to screen proteins involves a trade-off between efficiency and accuracy. To more effectively evaluate the functions of designed proteins, both oracle model-based and retrieval-based metrics have been proposed. 

\begin{figure*}[!htb]
    \centering
    \begin{subfigure}[b]{0.25\textwidth}
        \centering
        \includegraphics[width=\textwidth]{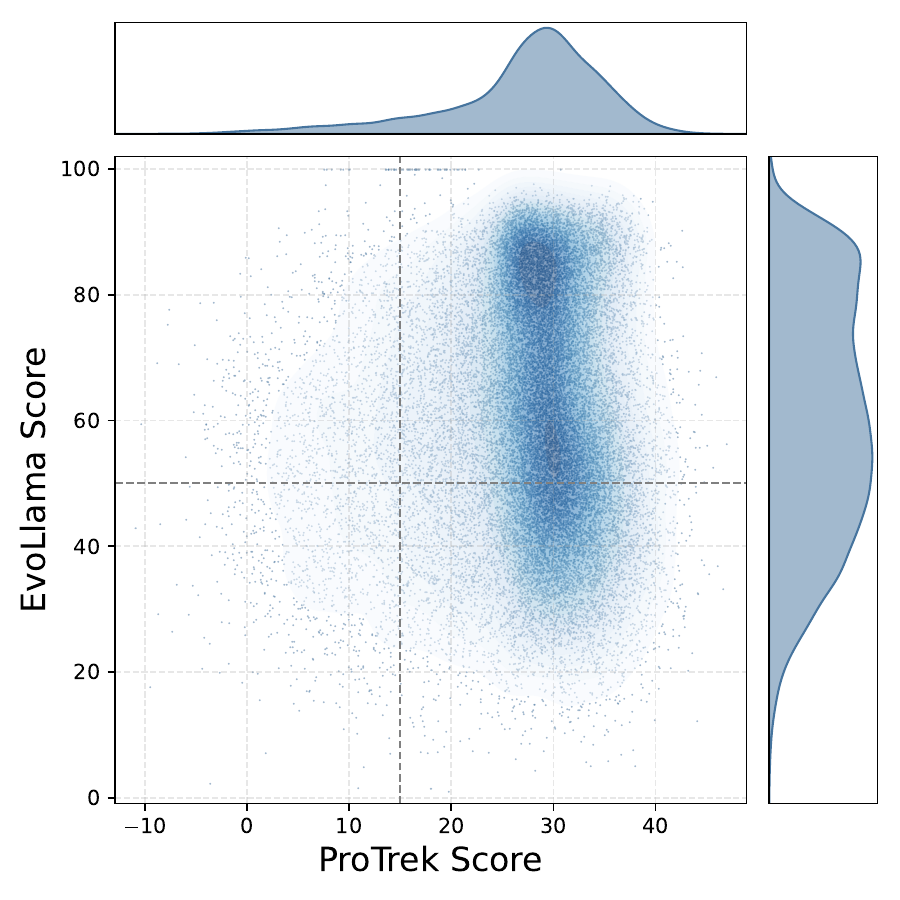}
        \caption{}
    \end{subfigure}
    \begin{subfigure}[b]{0.36\textwidth}
        \centering
        \includegraphics[width=\textwidth]{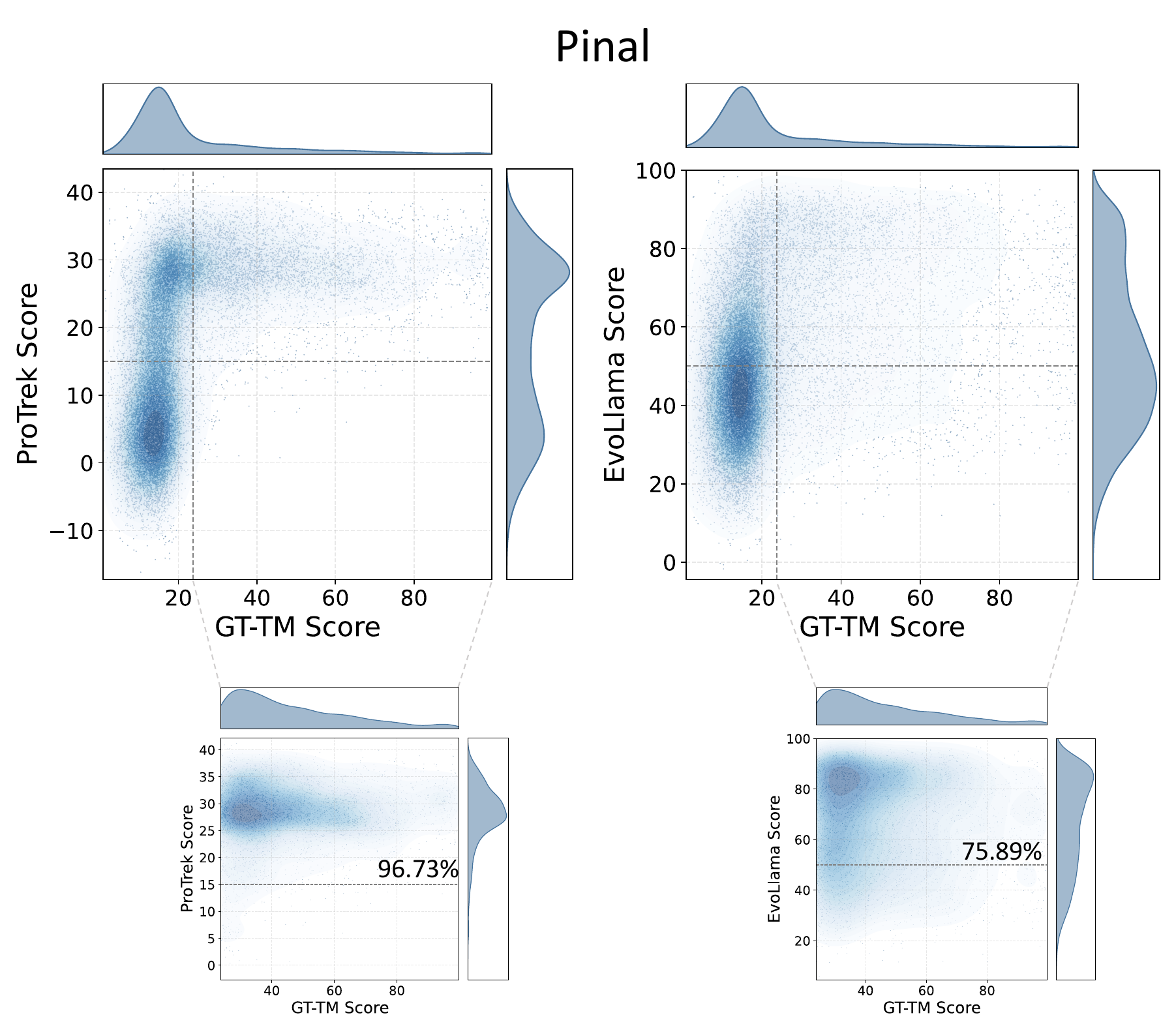}
        \caption{}
    \end{subfigure}
    \begin{subfigure}[b]{0.36\textwidth}
        \centering
        \includegraphics[width=\textwidth]{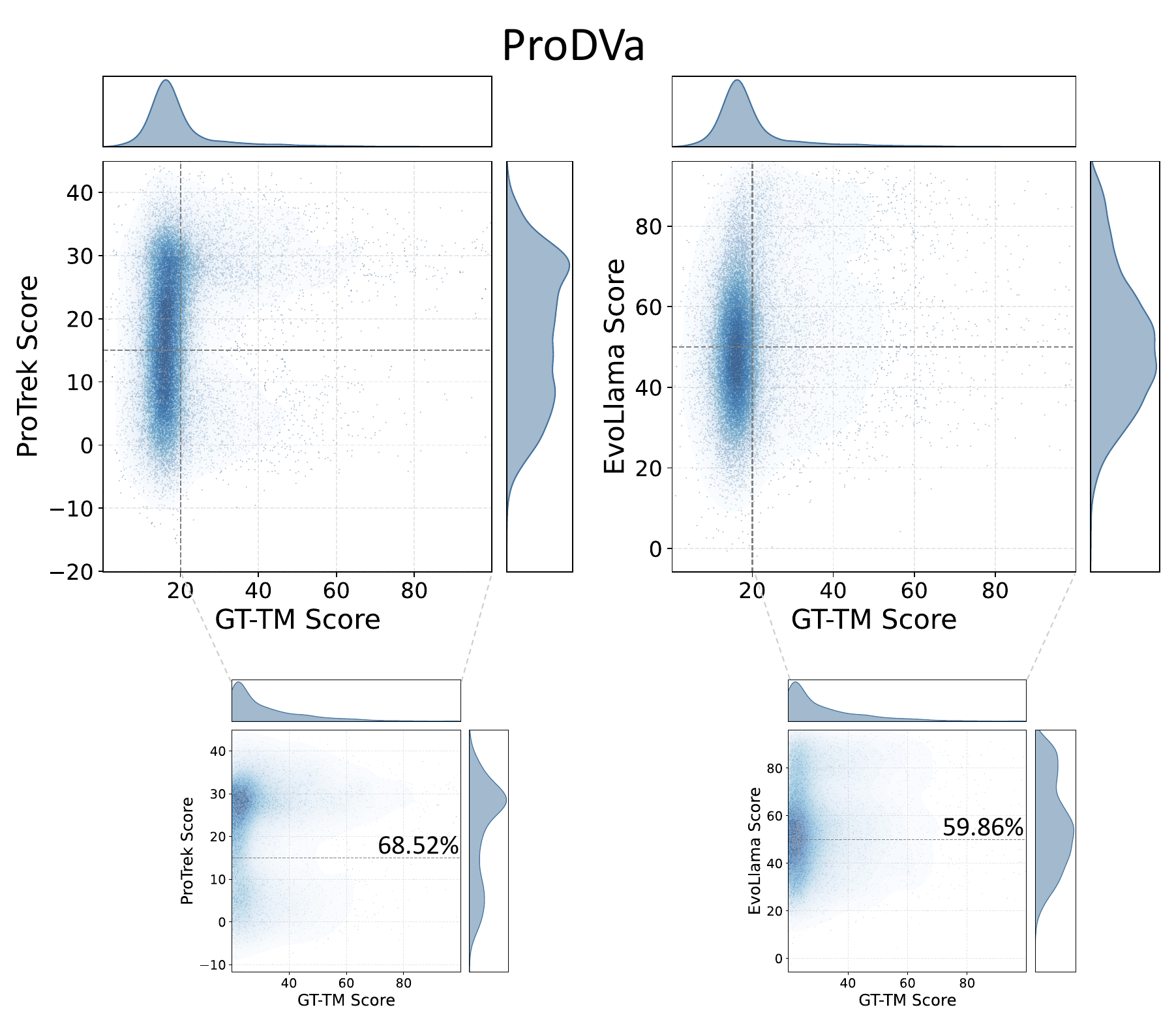}
        \caption{}
    \end{subfigure}
    \caption{\textbf{(a)} presents the distribution of the ProTrek Score and EvoLlama Score for natural proteins. \textbf{(b)} and \textbf{(c)} present the distributions of the GT-TM Score, ProTrek Score, and EvoLlama Score for proteins designed by Pinal and the ProDVa.}
    \label{fig:language-oracle}
\end{figure*}

We first investigate whether the two oracle model-based language alignment metrics exhibit consistency in evaluating natural proteins. These two metrics differ in two key perspectives. First, the ProTrek Score measures similarity between ground truth and designed proteins directly based on their embeddings, whereas the EvoLlama Score assesses similarity through predicted functional descriptions. Second, ProTrek is an oracle model pre-trained on large-scale datasets without further fine-tuning on specific downstream tasks. In contrast, EvoLlama is trained from scratch on the downstream task, leading to a distinct intrinsic knowledge distribution between the two oracle models. Figure \ref{fig:language-oracle}(a) illustrates the consistency between the ProTrek Score and the EvoLlama Score for natural proteins sampled from the validation and test sets of our description-guided task. The results show that most proteins are accurately predicted and tightly clustered in the upper-right corner, indicating strong agreement between the two metrics. Furthermore, following \citep{dai2024toward} and the definitions introduced in Section~\ref{sec:metric}, we establish empirical score thresholds to identify well-aligned proteins. Specifically, proteins with a ProTrek Score above 15 and an EvoLlama Score above 50 are considered to faithfully match the input functional descriptions.

\takeaway{The ProTrek Score and the EvoLlama Score are two oracle-based metrics that demonstrate high agreement in evaluating language alignment.}

The above discussion has remained focused on the global level of protein function. However, attention must also be directed toward local sequence alignment within proteins, particularly minor mutations in functional sub-sequences (motifs). To assess whether ProTrek is sensitive to protein mutations, we randomly select 1,000 natural proteins from \desctest and introduce random mutations with specified probabilities. The results are illustrated in Figure~\ref{fig:motif-sen}.

\takeaway{The ProTrek Score assesses both global alignment between entire proteins and their functions and local alignment between motifs and functions.}

Next, we discuss the language alignment metrics that do not rely on oracle models. The GT-TM Score measures the similarity between a designed protein and the ground truth structure. Since protein structure determines function, it is generally assumed that structurally similar proteins exhibit similar functions. However, we argue whether proteins with similar functions can fold into dissimilar structures. In Figures \ref{fig:language-oracle}(b) and (c), proteins designed by Pinal and ProDVa are sampled for illustration. The average score reported in Table~\ref{tab:main-res-desc} is used as the threshold to determine whether the GT-TM Score is considered high. It can be observed that 96.73\% of the Pinal-designed proteins with high similarity to the ground truth exhibit high ProTrek scores (above 15), while 75.89\% achieve high EvoLlama scores (above 50). A similar conclusion can be drawn from the ProDVa-designed proteins, demonstrating that high structural similarity leads to similar functions. Furthermore, for proteins with lower structural similarity, no correlation between the two similarities is observed.

\takeaway{A high GT-TM Score indicates strong structural similarity, but structural similarity is not necessary for good language alignment.}

\begin{figure*}[!htb]
    \centering
    \begin{subfigure}[b]{0.32\textwidth}
        \centering
        \includegraphics[width=\textwidth]{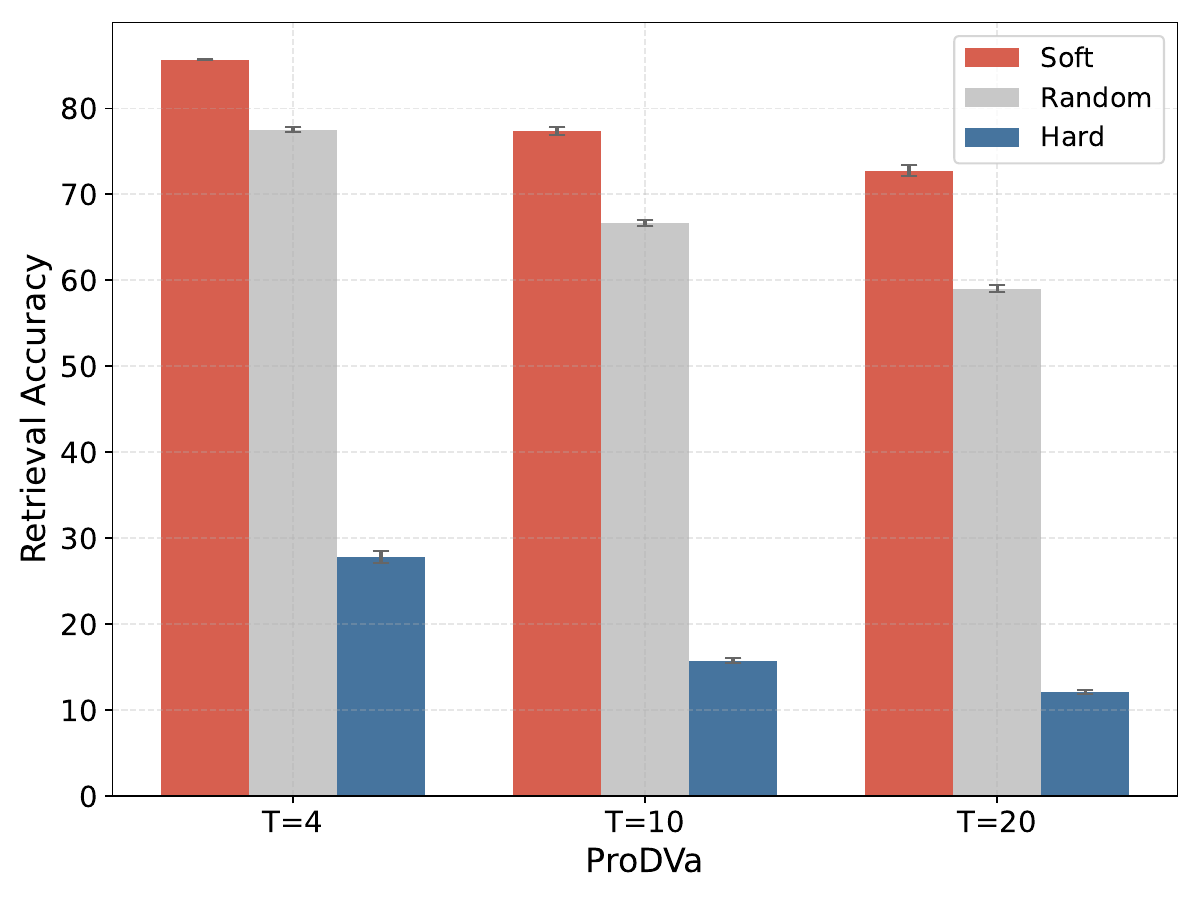}
        \caption{}
    \end{subfigure}
    \begin{subfigure}[b]{0.32\textwidth}
        \centering
        \includegraphics[width=\textwidth]{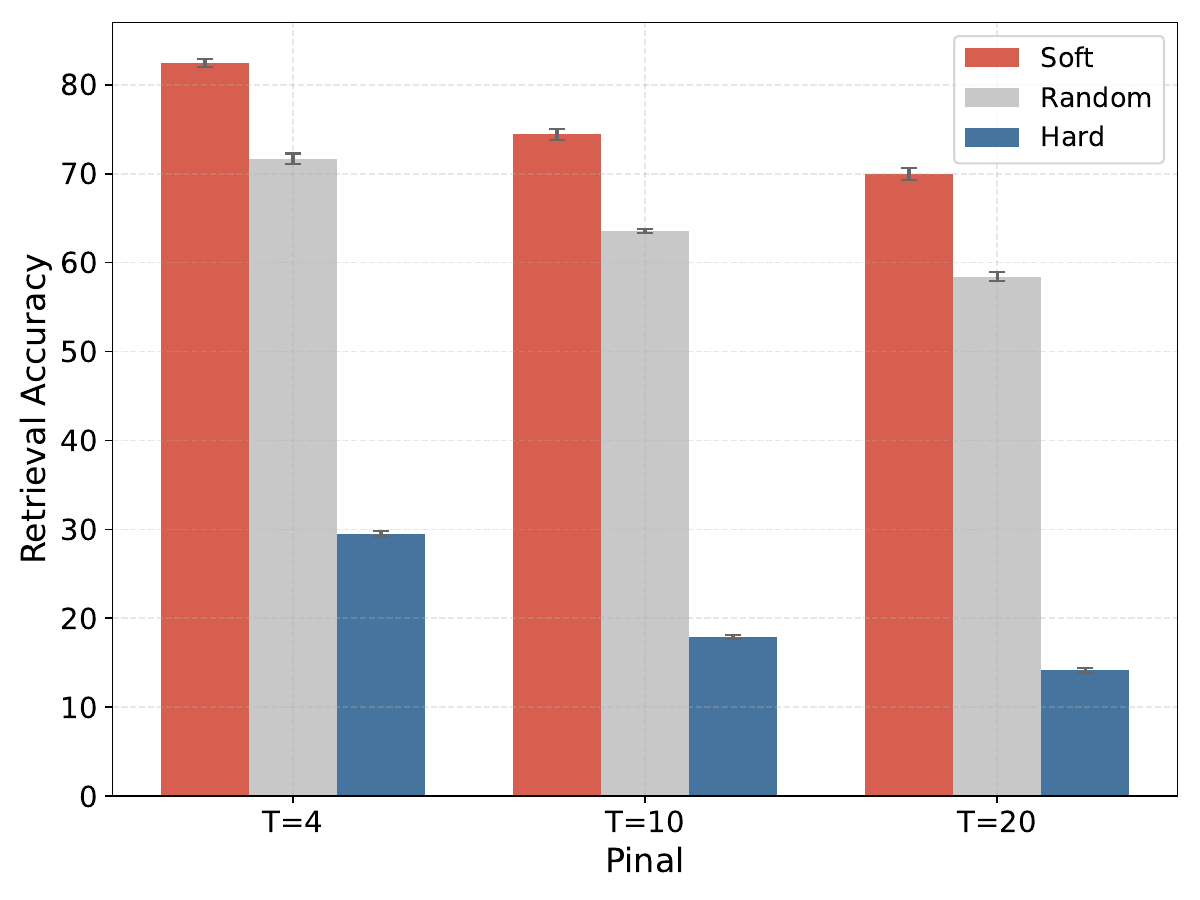}
        \caption{}
    \end{subfigure}
    \begin{subfigure}[b]{0.32\textwidth}
        \centering
        \includegraphics[width=\textwidth]{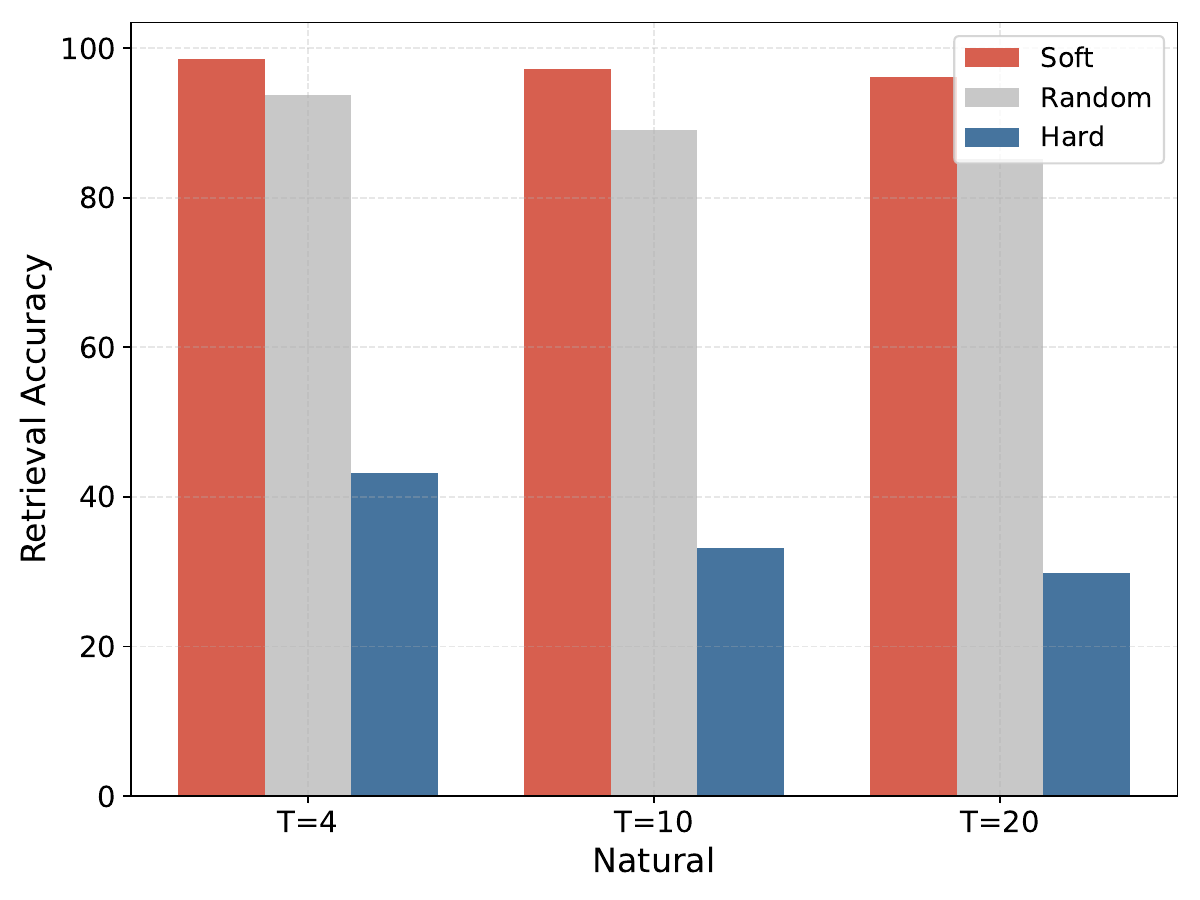}
        \caption{}
    \end{subfigure}
    \caption{Experimental results are reported for Soft, Random, and Hard Retrieval Accuracy. The random variant refers to the original implementation.}
    \label{fig:retrieval-accuracy}
\end{figure*}

In addition to the aforementioned metrics, Retrieval Accuracy is a retrieval-based metric that measures whether the embeddings of the positive function-sequence pair are the most similar among all candidates. However, this metric is highly dependent on the retrieved sequences. To assess the extent to which the retrieval strategy influences the results, we further define Soft Retrieval Accuracy and Hard Retrieval Accuracy. The difference between them lies in whether the \(T-1\) most or least relevant texts and their corresponding sequences are retrieved in relation to the positive pair. The relevance between textual descriptions is defined by the cosine similarity of their embeddings. In Figure \ref{fig:retrieval-accuracy}, for proteins designed by ProDVa and Pinal, the gap between Hard Retrieval Accuracy and Soft Retrieval Accuracy is 60.64\% and 55.82\%, respectively. Even for natural proteins, the gap between the two metrics on the ground truth can be as high as 66.31\%. Therefore, the results demonstrate that the retrieval strategy significantly impacts performance.

\takeaway{The retrieval strategy used in Retrieval Accuracy strongly affects the results. Random negative sampling can help, but the metric’s absolute values should be interpreted with caution.}

%% file: docs/020RelatedWork.tex
\Denovo protein design refers to the process of creating novel proteins from scratch, as opposed to the modification of existing sequences or structures~\citep{huang2016coming}. Current approaches in this field can be broadly categorized into unconditional and conditional generation methods. While unconditional generation operates without constraints, conditional generation methods—which guide model output using specific conditions—offer greater practical utility. Within the conditional generation domain, \ourbench primarily focuses on \denovo protein design from function. The detail of related work is presented in Appendix~\ref{sec:app-pdf}.

%% file: docs/070Conclusion.tex
The field of protein design has experienced growing interest in recent years, particularly in function-guided approaches. However, the lack of comprehensive and efficient evaluation benchmarks has hindered progress in this area. To address this gap, we introduce \ourbench, a benchmark designed to evaluate \denovo protein design from function. \ourbench focuses on four tasks and incorporates \mnum metrics to ensure a fair and comprehensive assessment. Additionally, we analyze the utility of these metrics and their interrelationships, offering deeper insights into \denovo protein design and the alignment between function and protein. Finally, we present best practices for using PDFBench in Appendix~\ref{sec:app-strategy}.

%% file: docs/080Appendix.tex
\section{Related Work}
\label{sec:app-pdf}
\input{docs/081RelatedWork.tex}

\section{Details of Datasets}
\label{sec:app-datasets}
\input{docs/082Datasets.tex}

\section{Details of Metrics}
\label{sec:app-metrics}
\input{docs/083Metrics.tex}

\section{Details of Baselines}
\label{sec:app-baselines}
\input{docs/084Baselines.tex}

\section{Details of Results}
\label{sec:app-results}
\input{docs/085Results.tex}

\section{Can Large Language Models Handle \Denovo Protein Design from Function?}
\label{sec:app-llm}
\input{docs/086LLM.tex}

\section{Best Practice for Benchmarking with PDFBench}
\label{sec:app-strategy}
\input{docs/087Practice.tex}

% \section{Differences from ProDVa}
% \label{sec:app-prodva}
% \input{docs/088ProDVa.tex}

\section{Limitations}
\label{sec:limitations}
\input{docs/088Limitation.tex}

%% file: docs/081RelatedWork.tex
\subsection{\Denovo Protein Design}
\Denovo protein design aims to generate novel protein sequences and can be categorized into unconditional and conditional approaches.

\paragraph{Unconditional Design} 
Unconditional design methods generate amino acid sequences without constraints, employing either autoregressive models~\citep{madani2023large,ferruz2022protgpt2,hesslow2022rita} or discrete diffusion models~\citep{alamdari2023protein,wang2024diffusion}. Alternatively, some methods~\citep{watson2023novo,mao2024de,wang2024proteus} adopt a two-stage paradigm: first generating backbone structures through diffusion in SE(3) space, then predicting corresponding sequences. Additionally, certain approaches~\citep{ren2024carbonnovo,wang2024dplm} utilize diffusion processes or energy-based models to jointly generate both backbone structures and sequences.

\paragraph{Conditional Design} 
Conditional \denovo design methods incorporate additional information to guide the design process, such as specified three-dimensional structures~\citep{harteveld2022deep,dauparas2022robust,hsu2022learning,gao2022pifold}, secondary structures~\citep{hu2024secondary}, protein-protein interactions~\citep{zhang2024ppi}, control tags~\citep{nijkamp2023progen2}, or function.

\subsection{\Denovo Protein Design from Function}

Within conditional \denovo, \denovo protein design from function (PDF) workflow has recently garnered substantial interest, existing methods can be categorized as description-guided methods and keyword-guided methods. 

\paragraph{Description-guided} 
description-guided methods include Chroma~\citep{ingraham2023illuminating}, ProteinDT~\citep{liu2025text}, PAAG~\citep{yuan2024annotation}, Pinal~\citep{dai2024toward}, ProDVa~\citep{liu2025proteindesigndynamicprotein}.
Chroma utilizes the diffusion framework to simultaneously design protein structure and sequence. 
ProteinDT employs a contrastive learning paradigm to align the representation spaces of function and protein, achieving over 90\% retrieval accuracy on its own evaluation. 
PAAG introduces a multi-level alignment module that enables simultaneous attention to both protein-level and domain-level information, resulting in superior performance on success rate metrics compared to previous approaches.  
Pinal differs from the end-to-end generation approaches mentioned above by first generating the protein structure based on function, followed by sequence design informed by both function and structure. Notably, Pinal utilizes the largest dataset, containing over 1000 times more protein-function pairs than all other methods. 
ProDVa employs a novel mechanism to dynamically retrieve the most relevant fragments from natural protein sequences based on the target function, significantly enhancing the method's performance.

\paragraph{Keyword-guided}
Keyword-guided methods include ProteoGAN~\citep{10.1093/bioinformatics/btac353}, ESM3~\citep{hayes2025simulating} and CFP-Gen~\citep{yin2025cfpgen}.
ProteoGAN employ a generative adversarial networks (GAN) to design protein sequences for 50 Gene Ontology terms.
ESM3, a cutting-edge multi-modal generative language model, integrates reasoning over protein sequence, structure, and function. It can respond to complex prompts combining these modalities and demonstrates high alignment responsiveness, enhancing its predictive fidelity. For keyword-guided task, it can generate sequences for 29026 IPR entries (nearly three-fifths of the InterPro).
CFP-Gen utilizes the latest generative framework-masked diffusion model-to design protein sequences for 375 GO terms and 1154 IPR entries.

As has been previously mentioned, all of the aforementioned methods claim to achieve optimal performance in their own reviews (i.e. their own proposed metrics). The objective of \ourbench is to provide a fair and comprehensive benchmark for evaluating the performance of the 8 novel methods.

\subsection{The Differences between PDF and the Other Workflows.} 

\paragraph{Differences from unconditional deisgn workflows.}
In unconditional design, the primary approach involves structure generation followed by inverse folding. For example, in the RFDiffusion+ProteinMPNN(RF+PM) workflow, RFDiffusion first generates a protein structure, and ProteinMPNN then designs the corresponding protein sequence based on this structure. The RF+PM workflow differs fundamentally from the PDF workflow in the following ways:

\begin{itemize}[leftmargin=*, itemsep=4pt, parsep=0pt, topsep=0pt]
    \item Input–Output: Although both pipelines generate protein sequences, the RF+PM workflow typically requires only gaussian noise or partial protein structures as input, whereas the PDF workflow uses natural language descriptions or keywords specifying protein functions.
    \item Application: Because RFDiffusion imposes input constraints, its workflow is typically applied to unconditional protein generation, motif scaffolding, binder design, and symmetric oligomer generation through the control of Gaussian noise features. In contrast, the PDF workflow has a broader range of applications.
\end{itemize}

\paragraph{Differences from the other conditional design workflows.}    
Compared with other conditional design methods, the PDF workflow provides greater flexibility and better aligns with practical requirements. It allows target protein specifications to be expressed directly in natural language rather than indirectly through secondary structure, protein interactions, or control tags. Furthermore, in practical protein design, target proteins often exhibit multi-level functional requirements (e.g., thermostability and catalytic activity). The PDF workflow naturally accommodates such complex functional demands.

\paragraph{Practical application value of PDF workflow.}
ProteinDT, Chroma, and ESM-3 have investigated the possibility of designing proteins from functions; however, none of these methods have been validated through wet-lab experiments.
Pinal presents 4 wet-lab experiments in its paper to determine whether in silico generative capabilities extend to producing biochemical functions. The study introduces a validation framework targeting four proteins of considerable scientific and industrial relevance. Each protein was selected to represent a distinct class of biochemical complexity: the intrinsic photophysics of GFP, the enzymatic catalysis of synthetic polymers by PETase, the cofactor-dependent mechanism of ADH, and the molecular coordination of an H-protein within a multi-protein system. The successful experimental realization of these four distinct protein classes offers definitive validation of the general applicability of this paradigm.
The PDF workflow does not merely accelerate existing research; it fundamentally changes the nature of protein design. Rather than being confined to rediscovering or marginally improving naturally occurring proteins, it enables the creation of entirely novel biological solutions.

%% file: docs/082Datasets.tex
\subsection{\desctest}
\label{sec:app-datasets-description}

Mol-Instructions~\citep{fang2023mol} comprises three key categories of instructions: molecule-oriented, protein-oriented, and biomolecular text. The protein-oriented instructions\footnote{\url{https://huggingface.co/datasets/zjunlp/Mol-Instructions}} include a 196K subset, Protein Design, for \denovo protein design, sourced from UniProtKB/Swiss-Prot and UniProtKB/TrEMBL. We use the test set of this subset with 5,876 proteins for our evaluation, referred to as \desctest.

\subsection{\keywtest}
\label{sec:app-datasets-keyword}

Unlike description-guided tasks, keyword-guided tasks lack a publicly available, high-quality evaluation dataset. To address this gap, we constructed a novel dataset. Specifically, we selected proteins from UniProt/SwissProt\footnote{\url{https://www.uniprot.org/uniprotkb?query=*}} released between January 1, 2025, and August 25, 2025, and collected their corresponding protein sequences, InterPro IDs, and Gene Ontology terms\footnote{Search with \texttt{(date\_created:[2025-01-01 TO 2025-08-25]) AND (reviewed:true)}}. To allow description-guided baselines to participate in the evaluation, we followed the approach of Mol-Instructions~\citep{fang2023mol} by concatenating the text descriptions associated with InterPro entries and Gene Ontology terms using a prompt. The detailed processing steps are provided in Table~\ref{tab:app-data-prompt}. We finally curated a novel dataset containing 1,057 proteins with 1,297 IPR entries and 380 GO terms, referred to as \keywtest.

\input{tabs/app/datasets/convert_template.tex}

\subsection{Fairness Analysis}
\label{sec:app-datasets-fairness}

For the description-guided task, the \desctest is excluded from the training process of all selected baseline models, including ProteinDT~\citep{liu2025text}, Chroma~\citep{ingraham2023illuminating}, PAAG~\citep{yuan2024annotation}, Pinal~\citep{dai2024toward}, and ProDVa~\citep{liu2025proteindesigndynamicprotein}. For the keyword-guided task, we apply a datetime cutoff and exclude all proteins and their corresponding keyword annotations dated prior to January 1, 2025, thereby strictly preventing potential data leakage. Therefore, there is no risk of data contamination (hard overlaps) during the evaluation.

While we also consider soft overlap settings (i.e., cases involving similar functions or sequences), in \ourbench, functions are provided as inputs, and the model is tasked with designing novel proteins (outputs) aligned with these functions. The model’s predictions are conditioned solely on the provided functions, and any overlap in outputs across splits neither confers an unfair advantage nor compromises the integrity of the evaluation. Therefore, the presence of similar proteins does not constitute data contamination in the conventional sense and has minimal impact on the evaluation.
For soft overlaps in similar functions, we conduct experiments to validate the fairness of our evaluation on the three baselines with publicly available training set, including ProDVa~\citep{liu2025proteindesigndynamicprotein}, ProteinDT~\citep{liu2025text}, and PAAG~\citep{yuan2024annotation}.
First, we use ProTrek~\citep{su2024protrek} to compute the embeddings of the functions and sequences in the three training sets and \desctest. Then, for each function–sequence pair in \desctest, if an identical sequence exists in the corresponding training set, we compute the cosine similarity between ProTrek embeddings of the function in \desctest and the corresponding function in the training set. If the similarity exceeds the \texttt{Threshold}, it is considered a soft overlap. We categorize the baseline results on \desctest into two groups based on the presence or absence of overlap, as shown in Table~\ref{tab:app-data-overlap}. The analysis shows that, for ProDVa and ProteinDT, excluding soft overlaps has a positive impact on Perplexity (up to 2.12\%), ProTrek Score (up to 2.74\%), and Retrieval Accuracy (up to 4.56\%). We observe minimal impact on foldability, despite slight improvements in pLDDT and PAE when soft overlaps are included. Additionally, for PAAG, the impact of including soft overlaps is consistently minimal across all metrics.

\input{tabs/app/datasets/overlap_analysis.tex}

%% file: tabs/app/datasets/convert_template.tex
\begin{table}[!htb]
    \caption{Prompt and templates for converting keywords into textual description.}
    \label{tab:app-data-prompt}
    \centering
    \resizebox{\textwidth}{!}{
    \renewcommand{\arraystretch}{1.8}
    \begin{tabular}{ll}
        \toprule
        \multicolumn{2}{l}{{\large \textbf{Prompt}}} \\
        \midrule
        \multicolumn{2}{l}{\makecell[c]{Generate a protein sequence for a novel protein that integrates the following function keywords:\\ \{textual annotations for all keywords connected by semicolons\}. The designed protein sequence is}}  \\
        \midrule
        {\large \textbf{Keyword Type}} & {\large \textbf{Template}} \\
        \midrule
        InterPro (Domain) & 
        The protein should contain one or more \{\texttt{domains}\} that are essential for its biological function \\
        InterPro (Family) & 
        The protein should belong to \{\texttt{family\(s\)}\} that shares evolutionary origin and functional similarity \\
        InterPro (Homologous\_Superfamily) &
        The protein should be classified within \{\texttt{homologous superfamily\(s\)}\} sharing conserved structural features \\
        InterPro (Repeat) &
         The protein should include one or more \{\texttt{repeat}\} that provide structural or functional support \\
        InterPro (Conserved\_Site) &
         The protein should contain \{\texttt{conversed site\(s\)}\} that is preserved across related proteins \\
        InterPro (Active\_Site) &
         The protein must have \{\texttt{activate sites}\} that is conserved among related catalytic enzymes \\
        InterPro (Binding\_Site) &
         The protein should include a \{\texttt{binding site\(s\)}\} that enables ligand binding under diverse conditions \\
        InterPro (PTM) &
         The protein should contain \{\texttt{PTM(s)}\} that allow regulation through chemical modifications \\
        Gene Ontology (Molecular Function) &
         The protein must be able to perform the \{molecular function\(s\)\} required for its activity \\
        \bottomrule
    \end{tabular}
    }
\end{table}

%% file: tabs/app/datasets/overlap_analysis.tex
\begin{table}[!htb]
    \caption{Soft overlap analysis on \desctest. Results including soft overlaps are presented in parentheses, while those excluding soft overlaps are shown outside the parentheses.}
    \label{tab:app-data-overlap}
    \centering
    \resizebox{\textwidth}{!}{
    \renewcommand{\arraystretch}{1.2}
    \begin{tabular}{lcccccc}
        \toprule
        \textbf{Models} & \textbf{\#Threshold\ (\#Num)} & 
        \textbf{Perplexity} \(\downarrow\) & \textbf{pLDDT} \(\uparrow\) & \textbf{PAE} \(\downarrow\) & \textbf{ProTrek Score} \(\uparrow\) & \textbf{Retrieval Accuracy} \(\uparrow\) \\
        \midrule
        \multirow{5}{*}{ProDVa} 
        & 0.0 (5876) & -\ (7.63) & -\ (76.86) & -\ (8.66) & -\ (17.40) & -\ (66.83) \\ 
        & 0.1 (1626) & 7.13\ (8.95) & 76.75\ (77.06) & 9.08\ (7.61) & 18.16\ (15.42) & 67.93\ (63.37) \\
        & 0.3 (185) & 7.56\ (9.68) & 76.81\ (77.61) & 8.72\ (7.42) & 17.49\ (14.78) & 66.83\ (61.62) \\
        & 0.5 (14) & 7.63\ (9.63) & 76.83\ (79.21) & 8.68\ (7.51) & 17.41\ (14.71) & 66.67\ (66.67) \\
        & 0.7 (2) & 7.63\ (5.54) & 76.84\ (78.23) & 8.68\ (5.67) & 17.40\ (23.80) & 66.66\ (100.00) \\
        \midrule
        \multirow{5}{*}{ProteinDT} 
        & 0.0 (5876) & -\ (12.41) & -\ (38.29) & -\ (25.13) & -\ (1.20) & -\ (16.92) \\ 
        & 0.1 (1524) & 12.45\ (12.29) & 37.50\ (40.55) & 25.12\ (25.17) & 1.06\ (1.59) & 16.35\ (17.15) \\
        & 0.3 (112) & 12.40\ (12.49) & 38.26\ (39.77) & 25.13\ (25.29) & 1.20\ (1.05) & 16.60\ (14.58) \\
        & 0.5 (5) & 12.41\ (11.55) & 38.29\ (44.35) & 25.13\ (26.21) & 1.20\ (1.96) & 16.56\ (13.33) \\
        & 0.7 (0) & - & - & - & - & - \\
        \midrule
        \multirow{5}{*}{PAAG} 
        & 0.0 (5876) & -\ (17.84) & -\ (28.39) & -\ (25.38) & -\ (1.29) & -\ (13.43 ) \\ 
        & 0.1 (749) & 17.85\ (17.79) & 28.37\ (28.53) & 25.36\ (25.47) & 1.18\ (2.00) & 12.65\ (14.11) \\
        & 0.3 (698) & 17.85\ (17.79) & 28.38\ (28.49) & 25.36\ (25.51) & 1.20\ (1.89) & 12.67\ (14.04) \\
        & 0.5 (579) & 17.85\ (17.79) & 28.40\ (28.29) & 25.36\ (25.55) & 1.24\ (1.74) & 12.67\ (14.28) \\
        & 0.7 (318) & 17.84\ (17.83) & 28.40\ (28.20) & 25.37\ (25.57) & 1.27\ (1.53) & 12.70\ (15.09) \\
        \bottomrule
    \end{tabular}
    }
\end{table}

%% file: docs/083Metrics.tex
% region Plausibility
\subsection{Plausibility}

\paragraph{Repetitiveness} 
We use the \textbf{RepN} metric to reflect the repetition at n-gram levels in the designed protein sequence \(P\). Additionally, we propose a metric, \textbf{Repeat}, to more accurately evaluate the proportion of repetitive sequence fragments from a biological perspective.

\begin{equation*}
    \text{Rep-n} = 100 \times (1.0 - \frac{\text{unique n-grams}(P)}{\text{total n-grams}(P)})
\end{equation*}

\begin{algorithm}
  \caption{Compute Repeat}
  \begin{algorithmic}
    \REQUIRE $sequence \neq \emptyset$
    \ENSURE $proportion \in [0.0, 1.0]$
    \STATE $n \gets sequence$
    \IF{$n = 0$}
      \STATE \textbf{return} $0.0$
    \ENDIF

    \STATE $regions \gets \emptyset$
    \STATE $max\_window\_size \gets \min\bigl(20,\lfloor n/2\rfloor\bigr)$

    \FOR{$window\_size = 1$ \textbf{to}\ $max\_window\_size$}
      \FOR{$i = 0$ \textbf{to}\ $n - window\_size$}
        \STATE $pattern \gets sequence[i:i+window\_size]$
        \STATE $count \gets 1$
        \STATE $j \gets i + window\_size$
        \WHILE{$j \le n - window\_size$ \algorithmicand\ sequence[j:j+window\_size] = pattern}
          \STATE $count \gets count + 1$
          \STATE $j \gets j + window\_size$
        \ENDWHILE
        \IF{$count \ge 3$}
          \STATE $regions \gets regions \cup \{(i,\; i + window\_size \times count)\}$
        \ENDIF
      \ENDFOR
    \ENDFOR

    \IF{$regions = \emptyset$}
      \STATE \textbf{return} $0.0$
    \ENDIF

    \STATE $merged \gets \text{sorted and merged regions}$
    \STATE $total\_repeat \gets \sum_{(start,end)\in merged} (end - start)$
    \STATE $proportion \gets total\_repeat / n$
    \STATE \textbf{return} $proportion$
  \end{algorithmic}
\end{algorithm}
% endregion 

% region Language Alignment
\subsection{Language Alignment}

\paragraph{ProTrek Score} 
 Given a designed sequence \(P\) and respective description \(T\), \textbf{ProTrek score} is defined as:

\begin{equation*}
    \text{ProTrek Score} = \cos(\tau_{\text{seq}}(P), \tau_{\text{text}}(T))/t
\end{equation*}

where\(\tau_{\text{seq}}(P)\), \(\tau_{\text{text}}(T)\) and \(t\) represent the protein sequence encoder, text encoder and temperature of ProtTrek, respectively.

\paragraph{EvoLlama Score} 
In addition to the ProTrek Score, we employ a generative approach for alignment evaluation using EvoLlama~\citep{liu2024evollama}. Specifically, we utilize EvoLlama, which comprises a 650M ESM2 protein sequence encoder and a 3B Llama-3.2 text decoder. The model is randomly initialized and trained from scratch using the SwissMolinst dataset described in \cite{liu2025proteindesigndynamicprotein}. Given the designed protein \(P\) and respective function description \(t\), we prompt the \(P\) with \texttt{The function of the protein is} and fetch EvoLlama-generated function description \(t'\) with it. Assume that the \(t\) and \(t'\) can be tokenized into \(k\) and \(k'\) tokens, respectively. The \textbf{EvoLlama Score} is defined as follows:
\begin{equation*}
  \text{EvoLlama Score} = \mathrm{sim}(\frac{1}{k} \sum_{i=1}^{k}\mathtt{Embed}(t), \frac{1}{k'} \sum_{i=1}^{k'}\mathtt{Embed}(t'))
\end{equation*}
where \(\mathtt{Embed}(\cdot)\) denotes using PubMedBERT \citep{gu2021domain} as the embedding model.

\paragraph{Keyword Recovery}
Given the designed sequence \(P\) and ground truth sequence \(GT\), \textbf{IPR Recovery} and \textbf{GO Recovery} are defined as:

\begin{equation*}
\text{IPR Recovery} =
\begin{cases}
\dfrac{\mathrm{InterProScan}(P) \cap \mathrm{InterProScanE}(GT)}{\mathrm{InterProScan}(GT)}, &
\quad\text{if } \mathrm{InterProScanO}(GT) \neq \emptyset,\\
\text{N/A}, &
\quad\text{if } \mathrm{InterProScan}(GT) = \emptyset. \\
\end{cases}
\end{equation*}

\begin{equation*}
\text{GO Recovery} =
\begin{cases}
\dfrac{\mathrm{DeepGO\text{-}SE}(P) \cap \mathrm{DeepGO\text{-}SE}(GT)}{\mathrm{DeepGO\text{-}SE}(GT)}, &
\quad\text{if } \mathrm{DeepGO\text{-}SE}(GT) \neq \emptyset,\\
\text{N/A}, &
\quad\text{if } \mathrm{DeepGO\text{-}SE}(GT) = \emptyset. \\
\end{cases}
\end{equation*}

\paragraph{Retrieval Accuracy} 
\textbf{Retrieval Accuracy} is obtained with a well-pretrained model, i.e. ProTrek. Given the designed sequence \(P\), respective description \(T_{0}\) and N randomly selected descriptions as negative pool \(\mathcal{N} = (T_1, T_2, ..., T_{N-1})\) from testing set, retrieval accuracy is defined as: 

\begin{equation*}
    \text{Retrieval Accuracy} = \left[ \cos(\tau_{\text{seq}}(P), \tau_{\text{text}}(T_0)) \ge \cos(\tau_{\text{seq}}(P), \tau_{\text{text}}(T_i)), \quad \forall i \in \mathcal{N} \right]
\end{equation*}

% endregion Language Alignment

% region Similarity
\subsection{Similarity}

\paragraph{GT-Identity}
Given the designed sequence \(P\) and ground truth sequence \(GT\), we compute the \textbf{GT-Identity} with the align module of MMseqs~\citep{kallenborn2024gpu}.

\begin{equation*}
    \text{GT-Identity} = \text{MMseqs\textsuperscript{align}}(P, GT)
\end{equation*}

\paragraph{ESMScore}
Following the formula in~\citep{bert-score}, we compute the BertScore between the ground truth sequence \(GT\) and the designed sequence \(P\) using ESM-2-650M~\citep{lin2022language}, namely \textbf{ESMScore}.

\paragraph{GT-TMscore}
We measures the \textbf{GT-TMscore} between the ESMFold-predicted structures between the design sequence \(P\) and the ground truth sequence \(GT\) using TMscore.

\begin{equation*}
    \text{GT-TMscore} = \text{TMscore}(\text{Structure}_{P}, \text{Structure}_{GT})
\end{equation*}
% endregion Similarity

% region Novelty
\subsection{Novelty}

\paragraph{Sequence Novelty}

Initially, MMseqs2 is utilized to retrieve the \(num\_prot\) most similar sequences and the respective similarity \(sim_i\) of the designed sequence \(P\) to UniProtKB. Subsequently, it is possible to obtain each novelty \(nov_i\) via \(1 - sim_i\) if a match sequence is present, otherwise \(1\). Finally, \textbf{Novelty-Seq\textsubscript{Easy}} and \textbf{Novelty-Seq\textsubscript{Hard}} can be defined as:

\begin{equation*}
    \text{Novelty-Seq}_{\text{Easy}}
    = \frac{\sum nov_i}{num\_prot},\ 
    \text{Novelty-Seq}_{\text{Hard}}
    = 1 - \max_{i} sim_i
\end{equation*}

\paragraph{Stucture Novelty}
Similar to the Sequence Novelty, given the designed sequence \(P\), Foldseek~\citep{kimRapidSensitiveProtein2025} is used to retrieve the \(num\_prot\) most similar structures and their respective similarities \(sim_i\) between the ESMFold-predicted structure \(Structure_P\) to AlphafoldDB/SwissProt\footnote{\url{https://alphafold.ebi.ac.uk/download\#swissprot-section}}.The values of \(nov_i\)are then obtained in the same manner as for Sequence Novelty. Finally, \textbf{Novelty-Struct\textsubscript{Easy}} and \textbf{Novelty-Struct\textsubscript{Hard}} can be defined as:

\begin{equation*}
   \text{Novelty-Struct}_{\text{Easy}}
    = \frac{\sum nov_i}{num_{prot}},\ 
    \text{Novelty-Struct}_{\text{Hard}}
    = 1 - \max_{i} sim_i
\end{equation*}

% region Novelty

\subsection{Diversity}
\paragraph{Sequence Diversity}
Given the \(N\) sequences \(\mathcal{P} = (P_1, P_2, ..., P_N)\) designed from the same description, we employ MMseqs2 to compute the similarity between each pair of sequences in \(\mathcal{P}\). The same as Novelty, we obtain each diversity \(div_i\) via \(1 - sim_i\) if the similarity between two sequences can be computed, otherwise \(1\), which means the two sequences are totally different. Finally, \textbf{Diversity-Seq} is defined as:

\begin{equation*}
    \text{Diversity-Seq} = \frac{\sum{div_i}}{N(N-1)}
\end{equation*}

\paragraph{Structure Diversity}
Given the \(N\) sequences \(\mathcal{P} = (P_1, P_2, ..., P_N)\) designed from the same description and their corresponding ESMFold-predicted structures \(\mathcal{S} = (S_1, S_2, ..., S_N)\), we use Foldseek to compute pairwise structural similarities within \(\mathcal{S}\), , yielding similarity scores \(sim_i\) and corresponding diversity values \(div_i\). Finally, \textbf{Diversity-Struct} is defined as:

\begin{equation*}
    \text{Diversity-Struct} = \frac{\sum{div_i}}{N(N-1)}
\end{equation*}

%% file: docs/084Baselines.tex
\subsection{Description-guided design baselines} 

\paragraph{Chroma} Chroma~\citep{ingraham2023illuminating} is a diffusion-based and programmable generation model for proteins. It employs a structured diffusion process that takes the physical properties of proteins as polymer chains into consideration. Chroma not only generates protein sequences using functional keywords, but it is also highly "programmable": users can guide the generation process using composable conditioners to enforce constraints such as symmetry, substructure, shape, or semantic properties (e.g., protein class, text prompts). 

\paragraph{ProteinDT} ProteinDT~\citep{liu2025text} is a multimodal framework that utilizes function descriptions to guide protein design. The fundamental principle underpinning the system is ProteinCLAP (Contrastive LAnguage and Protein), which employs comparative learning  to align description and protein sequence representations within a shared embedding space. The framework is composed of three sequential steps: ProteinCLAP aligns the representation, a facilitator maps the text embedding to the protein representation, and a autoregressive decoder generates a sequence based on this representation. 

\paragraph{PAAG} PAAG~\citep{yuan2024annotation} is a multimodal framework that focuses on both functional descriptions and functional keywords. It employs a multilevel alignment module that aligns protein sequences and description/keywords at the global and local levels. Subsequently, an autoregressive decoder is employed to generate protein sequences based on the aligned annotation representations. 

\paragraph{Pinal} Pianl~\citep{dai2024toward} is a large-scale(up to 16B parameters) \denovo protein design framework intended to translate natural language instructions into novel protein sequences. In lieu of direct text-to-sequence generation, Pinal adopts a two-stage approach: first, protein structures are generated from linguistic descriptions; then, sequences are designed based on the generated structures and the original linguistic input.  This strategy employs the relatively restricted structure space as a preliminary step to efficiently constrain the extensive sequence search space. 

\paragraph{ProDVa} ProDVa~\citep{liu2025proteindesigndynamicprotein} is a multimodal protein design framework that combines textual function descriptions with insights from natural protein fragments to create sequences that are both functionally aligned and structurally plausible. It integrates a text encoder, a protein language model, and a fragment encoder that dynamically retrieves the most relevant fragments based on the desired function.

\subsection{Keyword-guided design baselines} 

\paragraph{ProteoGAN} ProteoGAN~\citep{10.1093/bioinformatics/btac353} is a conditional generative adversarial network designed to generate novel protein sequences based on functional labels from the Gene Ontology (GO).

\paragraph{ESM3} ESM3~\citep{hayes2025simulating} is a large-scale (up to 98 billion parameters) multimodal generative language model designed to simulate protein evolution. It represents protein sequence, structure, and function as discrete tokens processed within a unified latent space. ESM3 is trained using a generative masked language modeling (MLM) objective, predicting randomly masked tokens across modalities to learn their complex interrelationships. The model can follow complex prompts combining sequence, structure, and function information. 

\paragraph{CFP-Gen} CFP-Gen~\citep{yin2025cfpgen} is a novel diffusion language model designed for generating functional proteins by simultaneously integrating multiple constraints from different modalities, including function, sequence, and structure. It employs an Annotation-Guided Feature Modulation (AGFM) module to control protein features using functional annotations and a Residue-Controlled Functional Encoding (RCFE) module for precise residue-level control.

%% file: docs/085Results.tex
\subsection{Experiment Setting}
\label{sec:app-res-set}
We implement the baselines as follows,

\noindent \textbf{Description-guided}
For description-guided baselines (ProDVa, Pinal, PAAG, Chroma, ProteinDT), we directly use the description as input to prompt the model to generate sequences.

\noindent \textbf{Keyword-guided} 
Keyword-guided baselines only support restricted keywords, necessitating additional data processing. For InterPro entries unsupported by ESM3 and CFP-Gen, we opted to skip these keywords to ensure comparative fairness. For Gene Ontology terms unsupported by ProteoGAN and CFP-Gen, we attempted to find their ancestor terms in the GO database for substitution; if none were found, we skipped the term.

\noindent \textbf{All baselines}  We evaluated using the official baseline implementation and weights provided. The results are averaged across 3 runs with different random seeds to ensure fairness. Additionally, no further fine-tuning is performed on any of the baselines to ensure a fair evaluation of the models’ capability to design novel and functional proteins under the same settings.

\subsection{Complete Results for the Description-guided Task}
\label{sec:app-res-desc}
Results on keyword-guided task are in Table \ref{tab:app-res-desc-seq-struct}, Table \ref{tab:app-res-desc-align} and Table \ref{tab:app-res-desc-sim-nov-div}.  We mark the top three models, with deeper colors indicating superior performance.

\input{tabs/app/results/desc/seq_struct.tex}

\input{tabs/app/results/desc/alignment.tex}

\input{tabs/app/results/desc/sim_nov_div.tex}

\subsection{Complete Results for the Keyword-guided Task}
\label{sec:app-res-keyw}
Results on keyword-guided task are in Table \ref{tab:app-res-keyw-seq-struct}, Table \ref{tab:app-res-keyw-align} and Table \ref{tab:app-res-keyw-sim-nov-div}. We mark the top three models, with deeper colors indicating superior performance.

\input{tabs/app/results/keyw/seq_struct.tex}

\input{tabs/app/results/keyw/alignment.tex}

\input{tabs/app/results/keyw/sim_nov_div.tex}

%% file: tabs/app/results/desc/seq_struct.tex
\begin{table}[!htb]
    \caption{Benchmark results of Plausibility and foldability for the description-guided task.}
    \label{tab:app-res-desc-seq-struct}
    \centering
    \resizebox{\textwidth}{!}{
    \begin{tabular}{lcccccccccc}
        \toprule
        \multirow{2}{*}[-0.5ex]{Models} & \multicolumn{3}{c}{Perplexity} & \multicolumn{3}{c}{Repetitveness} & \multicolumn{4}{c}{Foldability}  \\
        \cmidrule[0.5pt](lr){2-4} \cmidrule[0.5pt](lr){5-7} \cmidrule[0.5pt](lr){8-11}
        & \makecell{PPL-ProtGPT2 \(\downarrow\)} 
        & \makecell{PPL-ProGen \(\downarrow\)}
        & \makecell{PPL-RITA \(\downarrow\)} 
        & \makecell{Repeat \(\downarrow\)} 
        & \makecell{Rep-2 \(\downarrow\)} 
        & \makecell{Rep-5 \(\downarrow\)} 
        & \makecell{pLDDT \(\uparrow\)} 
        & \makecell{\% > 70 \(\uparrow\)} 
        & \makecell{PAE \(\downarrow\)} 
        & \makecell{\% < 10 \(\uparrow\)}  \\ 
        \midrule
        Natural & 318.15 & 5.99 & 5.52 & 1.99 & 44.49 & 0.25 & 80.64 & 81.16 & 9.20 & 65.64 \\
        \midrule
        Random(U) & 2484.04\(\pm\)4.53 & 21.71\(\pm\)0.00 & 22.14\(\pm\)0.01 & 0.72\(\pm\)0.01 & 34.59\(\pm\)0.03 & 0.01\(\pm\)0.00 & 22.96\(\pm\)0.04 & 0.16\(\pm\)0.04 & 24.85\(\pm\)0.01 & 0.56\(\pm\)0.03 \\
        Random(E) & 3136.88\(\pm\)4.17 & 18.68\(\pm\)0.00 & 19.04\(\pm\)0.00 & 1.15\(\pm\)0.01 & 40.99\(\pm\)0.01 & 0.01\(\pm\)0.00 & 25.77\(\pm\)0.03 & 0.19\(\pm\)0.06 & 24.71\(\pm\)0.01 & 0.60\(\pm\)0.03 \\
        \midrule
        ProteinDT & 1576.23\(\pm\)4.32 & 12.41\(\pm\)0.01 & 12.44\(\pm\)0.01 & 6.83\(\pm\)0.10 & 62.47\(\pm\)0.14 & 2.82\(\pm\)0.05 & 38.29\(\pm\)0.04 & 0.98\(\pm\)0.17 & 25.13\(\pm\)0.02 & 0.40\(\pm\)0.09 \\
        Chroma & \third{1370.21\(\pm\)1.48} & \third{12.19\(\pm\)0.00} & \third{12.42\(\pm\)0.01} & \third{2.59\(\pm\)0.02} & \third{55.41\(\pm\)0.03} & \second{0.60\(\pm\)0.01} & \third{59.18\(\pm\)0.09} & \third{20.17\(\pm\)0.23} & \third{15.03\(\pm\)0.04} & \third{28.62\(\pm\)0.62} \\
        PAAG & 2782.70\(\pm\)9.63 & 17.84\(\pm\)0.01 & 18.05\(\pm\)0.02 & \second{2.34\(\pm\)0.02} & \second{45.83\(\pm\)0.03} & \first{0.09\(\pm\)0.01} & 28.39\(\pm\)0.07 & 0.07\(\pm\)0.03 & 25.38\(\pm\)0.01 & 0.10\(\pm\)0.03 \\
        Pinal & \first{308.97\(\pm\)0.68} & \first{5.81\(\pm\)0.02} & \first{5.78\(\pm\)0.02} & 12.83\(\pm\)0.13 & 58.26\(\pm\)0.16 & 4.73\(\pm\)0.06 & \second{75.25\(\pm\)0.19} & \second{68.93\(\pm\)0.33} & \second{10.96\(\pm\)0.10} & \second{58.41\(\pm\)0.38} \\
        ProDVa & \second{415.64\(\pm\)7.40} & \second{7.63\(\pm\)0.09} & \second{8.83\(\pm\)0.17} & \first{1.92\(\pm\)0.05} & \first{35.65\(\pm\)0.15} & \third{2.81\(\pm\)0.13} & \first{76.84\(\pm\)0.17} & \first{76.27\(\pm\)0.59} & \first{8.67\(\pm\)0.05} & \first{67.65\(\pm\)0.43} \\
        \bottomrule
    \end{tabular}
    }
\end{table}

%% file: tabs/app/results/desc/alignment.tex
\begin{table}[!htb]
    \caption{Benchmark results of Language Alignment for the description-guided task.}
    \label{tab:app-res-desc-align}
    \centering
    \resizebox{\textwidth}{!}{
    \begin{tabular}{lccccccccccccc}
        \toprule
        \multirow{2}{*}[-0.5ex]{Models} & \multicolumn{2}{c}{Model-based Alignment} & \multicolumn{9}{c}{Retrieval-based Alignment} \\
        \cmidrule[0.5pt](lr){2-3} \cmidrule[0.5pt](lr){4-12}
        & \makecell{ProTrek Score \(\uparrow\)} 
        & \makecell{EvoLlama Score \(\uparrow\)}
        & \makecell{Soft(4) \(\uparrow\)} 
        & \makecell{Soft(10) \(\uparrow\)} 
        & \makecell{Soft(20) \(\uparrow\)} 
        & \makecell{Normal(4) \(\uparrow\)} 
        & \makecell{Normal(10) \(\uparrow\)} 
        & \makecell{Normal(20) \(\uparrow\)} 
        & \makecell{Hard(4) \(\uparrow\)} 
        & \makecell{Hard(10) \(\uparrow\)} 
        & \makecell{Hard(20) \(\uparrow\)} \\ 
        \midrule
        Natural & 27.00 & 60.33 & 98.50 & 97.17 & 96.09 & 93.72 & 89.01 & 85.11 & 43.23 & 33.20 & 29.78 \\
        \midrule
        Random(U) & 1.03\(\pm\)0.04 & 36.22\(\pm\)0.07 & 28.09\(\pm\)0.83 & 12.62\(\pm\)0.29 & 6.94\(\pm\)0.20 & 28.97\(\pm\)0.38 & 12.83\(\pm\)0.52 & 7.16\(\pm\)0.29 & 25.95\(\pm\)0.52 & 10.57\(\pm\)0.40 & 5.38\(\pm\)0.39 \\
        Random(E) & 1.04\(\pm\)0.05 & 34.11\(\pm\)0.10 & 28.35\(\pm\)0.62 & 12.83\(\pm\)0.51 & 6.73\(\pm\)0.66 & 28.97\(\pm\)0.62 & 12.59\(\pm\)0.29 & 6.84\(\pm\)0.39 & 25.79\(\pm\)0.45 & 10.39\(\pm\)0.67 & 5.46\(\pm\)0.45 \\
        \midrule
        ProteinDT & 1.20\(\pm\)0.06 & \third{40.57\(\pm\)0.05} & \third{42.91\(\pm\)0.68} & \third{24.97\(\pm\)1.33} & \third{16.77\(\pm\)1.16} & \third{34.58\(\pm\)0.99} & \third{16.56\(\pm\)0.44} & \third{9.43\(\pm\)0.33} & 25.09\(\pm\)1.17 & 10.38\(\pm\)0.55 & 5.01\(\pm\)0.43 \\
        Chroma & \third{2.10\(\pm\)0.02} & 40.10\(\pm\)0.23 & 29.54\(\pm\)0.59 & 13.43\(\pm\)0.18 & 7.41\(\pm\)0.22 & 29.63\(\pm\)0.58 & 13.26\(\pm\)0.50 & 7.44\(\pm\)0.21 & \third{25.51\(\pm\)0.47} & \third{10.68\(\pm\)0.43} & \third{5.73\(\pm\)0.19} \\
        PAAG & 1.29\(\pm\)0.04 & 34.39\(\pm\)0.18 & 33.33\(\pm\)0.25 & 15.24\(\pm\)0.33 & 8.27\(\pm\)0.20 & 29.63\(\pm\)0.70 & 12.83\(\pm\)0.14 & 6.87\(\pm\)0.17 & 25.19\(\pm\)0.65 & 10.13\(\pm\)0.34 & 4.96\(\pm\)0.23 \\
        Pinal & \first{17.50\(\pm\)0.09} & \first{53.40\(\pm\)0.31} & \second{82.42\(\pm\)0.45} & \second{74.44\(\pm\)0.63} & \second{69.99\(\pm\)0.67} & \second{71.69\(\pm\)0.59} & \second{63.53\(\pm\)0.24} & \second{58.43\(\pm\)0.48} & \first{29.51\(\pm\)0.31} & \first{17.89\(\pm\)0.26} & \first{14.17\(\pm\)0.27} \\
        ProDVa & \second{17.40\(\pm\)0.06} & \second{51.19\(\pm\)0.17} & \first{85.64\(\pm\)0.06} & \first{77.37\(\pm\)0.44} & \first{72.75\(\pm\)0.61} & \first{77.52\(\pm\)0.28} & \first{66.67\(\pm\)0.37} & \first{59.03\(\pm\)0.44} & \second{27.84\(\pm\)0.71} & \second{15.77\(\pm\)0.29} & \second{12.11\(\pm\)0.28} \\
        \bottomrule
    \end{tabular}
    }
\end{table}

%% file: tabs/app/results/desc/sim_nov_div.tex
\begin{table}[!htb]
    \caption{Benchmark results of Similarity, Novelty, and Diversity for the description-guided task.}
    \label{tab:app-res-desc-sim-nov-div}
    \centering
    \resizebox{\textwidth}{!}{
    \begin{tabular}{lccccccccccc}
        \toprule
        \multirow{2}{*}[-0.5ex]{Models} & \multicolumn{5}{c}{Similarity} & \multicolumn{4}{c}{Novelty} & \multicolumn{2}{c}{Diversity}\\
        \cmidrule[0.5pt](lr){2-6} \cmidrule[0.5pt](lr){7-10} \cmidrule[0.5pt](lr){11-12}
        & \makecell{GT-Identity \(\uparrow\)} 
        & \makecell{GT-TMScore \(\uparrow\)}
        & \makecell{ESM-F1 \(\uparrow\)} 
        & \makecell{ESM-Precision \(\uparrow\)} 
        & \makecell{ESM-Recall \(\uparrow\)} 
        & \makecell{Seq\textsubscript{Easy} \(\uparrow\)} 
        & \makecell{Seq\textsubscript{Hard} \(\uparrow\)} 
        & \makecell{Struct\textsubscript{Easy} \(\uparrow\)} 
        & \makecell{Struct\textsubscript{Hard} \(\uparrow\)} 
        & \makecell{Seq \(\uparrow\)} 
        & \makecell{Struct \(\uparrow\)} \\ 
        \midrule
        Natural & 100.00 & 100.00 & 100.00 & 100.00 & 100.00 & 36.11 & 4.90 & 38.51 & 13.56 & - & - \\
        \midrule
        Random(U) & 0.37\(\pm\)0.03 & 16.95\(\pm\)0.03 & 71.06\(\pm\)0.02 & 81.66\(\pm\)0.02 & 63.46\(\pm\)0.02 & 98.77\(\pm\)0.03 & 58.14\(\pm\)0.07 & 96.82\(\pm\)0.03 & 77.64\(\pm\)0.12 & 98.61 & 81.59 \\
        Random(E) & 0.23\(\pm\)0.04 & 17.10\(\pm\)0.00 & 71.95\(\pm\)0.02 & 82.51\(\pm\)0.02 & 64.35\(\pm\)0.02 & 98.45\(\pm\)0.01 & 60.19\(\pm\)0.14 & 96.25\(\pm\)0.04 & 76.82\(\pm\)0.10 & 99.74 & 81.45 \\
        \midrule
        ProteinDT & 0.18\(\pm\)0.02 & 13.94\(\pm\)0.03 & 72.80\(\pm\)0.05 & \second{81.44\(\pm\)0.03} & 66.38\(\pm\)0.05 & \third{96.92\(\pm\)0.12} & \first{70.74\(\pm\)0.07} & \second{94.68\(\pm\)0.02} & \second{71.16\(\pm\)0.08} & \third{99.31} & \first{83.67} \\
        Chroma & \third{0.22\(\pm\)0.04} & \third{17.93\(\pm\)0.02} & 72.82\(\pm\)0.02 & \third{80.22\(\pm\)0.03} & \third{67.06\(\pm\)0.01} & \second{97.28\(\pm\)0.02} & \third{58.68\(\pm\)0.09} & \third{80.99\(\pm\)0.04} & \third{51.06\(\pm\)0.21} & \first{98.42} & \third{79.90} \\
        PAAG & 0.17\(\pm\)0.02 & 14.63\(\pm\)0.03 & \third{73.26\(\pm\)0.03} & \first{83.10\(\pm\)0.02} & 66.04\(\pm\)0.03 & \first{98.90\(\pm\)0.02} & \second{63.64\(\pm\)0.09} & \first{96.44\(\pm\)0.03} & \first{77.34\(\pm\)0.12} & \second{99.41} & \second{82.16} \\
        Pinal & \second{18.65\(\pm\)0.15} & \first{23.75\(\pm\)0.14} & \first{76.63\(\pm\)0.06} & 77.74\(\pm\)0.08 & \first{75.99\(\pm\)0.06} & 55.55\(\pm\)0.19 & 43.82\(\pm\)0.22 & 40.07\(\pm\)0.33 & 17.23\(\pm\)0.23 & 75.88 & 72.73 \\
        ProDVa & \first{21.48\(\pm\)0.15} & \second{20.03\(\pm\)0.11} & \second{75.23\(\pm\)0.01} & 77.01\(\pm\)0.05 & \second{74.11\(\pm\)0.02} & 38.23\(\pm\)0.31 & 14.64\(\pm\)0.23 & 56.18\(\pm\)23.36 & 36.31\(\pm\)33.02 & 46.10 & 52.59 \\
        \bottomrule
    \end{tabular}
    }
\end{table}

%% file: tabs/app/results/keyw/seq_struct.tex
\begin{table}[!htb]
    \caption{Benchmark results of Plausibility and Foldability for the keyword-guided task.}
    \label{tab:app-res-keyw-seq-struct}
    \centering
    \resizebox{\textwidth}{!}{
    \begin{tabular}{lccccccccccc}
        \toprule
        \multirow{2}{*}[-0.5ex]{Models} & \multicolumn{3}{c}{Perplexity} & \multicolumn{3}{c}{Repetitveness} & \multicolumn{4}{c}{Foldability}  \\
        \cmidrule[0.5pt](lr){2-4} \cmidrule[0.5pt](lr){5-7} \cmidrule[0.5pt](lr){8-11}
        & \makecell{PPL-ProtGPT2 \(\downarrow\)} 
        & \makecell{PPL-ProGen \(\downarrow\)}
        & \makecell{PPL-RITA \(\downarrow\)} 
        & \makecell{Repeat \(\downarrow\)} 
        & \makecell{Rep-2 \(\downarrow\)} 
        & \makecell{Rep-5 \(\downarrow\)} 
        & \makecell{pLDDT \(\uparrow\)} 
        & \makecell{\% > 70 \(\uparrow\)} 
        & \makecell{PAE \(\downarrow\)} 
        & \makecell{\% < 10 \(\uparrow\)}  \\  
        \midrule
        \multicolumn{11}{c}{\textbf{\textit{guided with GO keywords}}} \\
        \midrule
        \arrayrulecolor{gray!20}
        Natural & 554.35 & 9.17 & 8.89 & 2.17 & 44.43 & 0.43 & 76.92 & 72.44 & 10.54 & 54.69 \\
        \midrule
        Random(U) & 2473.84\(\pm\)10.48 & 21.74\(\pm\)0.01 & 22.18\(\pm\)0.02 & 0.72\(\pm\)0.03 & 35.52\(\pm\)0.04 & 0.01\(\pm\)0.00 & 23.20\(\pm\)0.02 & 0.10\(\pm\)0.08 & 24.56\(\pm\)0.01 & 0.19\(\pm\)0.08 \\
        Random(E) & 3096.13\(\pm\)18.42 & 18.68\(\pm\)0.01 & 19.05\(\pm\)0.01 & 1.14\(\pm\)0.04 & 41.45\(\pm\)0.07 & 0.02\(\pm\)0.00 & 25.99\(\pm\)0.12 & 0.05\(\pm\)0.08 & 24.47\(\pm\)0.03 & 0.24\(\pm\)0.08 \\
        \midrule
        GPT-5 & 1212.05\(\pm\)20.30 & 13.90\(\pm\)0.12 & 13.97\(\pm\)0.12 & 2.35\(\pm\)0.08 & 54.17\(\pm\)0.37 & 3.72\(\pm\)0.20 & 32.28\(\pm\)0.31 & 1.73\(\pm\)0.43 & 24.09\(\pm\)0.07 & 0.63\(\pm\)0.33 \\
        \midrule
        ProteoGAN & 2708.39\(\pm\)32.50 & 18.03\(\pm\)0.01 & 18.31\(\pm\)0.02 & \third{2.50\(\pm\)0.05} & \third{42.73\(\pm\)0.86} & \first{0.03\(\pm\)0.00} & 28.72\(\pm\)0.43 & 0.06\(\pm\)0.10 & 24.67\(\pm\)0.17 & 0.12\(\pm\)0.20 \\
        CFP-Gen & \first{187.72\(\pm\)9.71} & \first{5.16\(\pm\)0.03} & \first{4.65\(\pm\)0.02} & 12.67\(\pm\)0.79 & 59.67\(\pm\)0.83 & 13.82\(\pm\)0.74 & \second{73.38\(\pm\)0.26} & \second{65.65\(\pm\)1.11} & \third{14.61\(\pm\)0.27} & \third{35.20\(\pm\)1.76} \\
        ProteinDT & 1531.76\(\pm\)17.19 & 12.23\(\pm\)0.06 & \third{12.29\(\pm\)0.06} & 7.98\(\pm\)0.51 & 64.01\(\pm\)0.25 & 3.32\(\pm\)0.38 & 40.35\(\pm\)0.30 & 1.15\(\pm\)0.00 & 25.57\(\pm\)0.03 & 0.00\(\pm\)0.00 \\
        Chroma & 1354.61\(\pm\)4.81 & 12.18\(\pm\)0.03 & 12.40\(\pm\)0.03 & 2.71\(\pm\)0.05 & 55.09\(\pm\)0.12 & \third{0.67\(\pm\)0.03} & 59.27\(\pm\)0.20 & 22.17\(\pm\)0.65 & 15.00\(\pm\)0.10 & 30.93\(\pm\)0.08 \\
        PAAG & 2650.36\(\pm\)11.01 & 18.08\(\pm\)0.02 & 18.38\(\pm\)0.02 & \second{2.48\(\pm\)0.20} & \second{39.23\(\pm\)0.05} & \second{0.05\(\pm\)0.01} & 31.47\(\pm\)0.10 & 0.34\(\pm\)0.36 & 23.88\(\pm\)0.05 & 0.24\(\pm\)0.08 \\
        Pinal & \second{414.26\(\pm\)77.15} & \second{6.85\(\pm\)0.59} & \second{6.89\(\pm\)0.64} & 14.13\(\pm\)2.58 & 59.84\(\pm\)4.37 & 4.85\(\pm\)1.63 & \third{72.58\(\pm\)5.55} & \third{62.10\(\pm\)14.24} & \second{11.79\(\pm\)2.52} & \second{52.19\(\pm\)12.03} \\
        ProDVa & \third{486.77\(\pm\)9.51} & \third{11.16\(\pm\)0.29} & 18.71\(\pm\)0.77 & \first{1.87\(\pm\)0.07} & \first{22.04\(\pm\)0.09} & 0.88\(\pm\)0.05 & \first{74.73\(\pm\)0.24} & \first{68.40\(\pm\)0.38} & \first{6.11\(\pm\)0.02} & \first{84.90\(\pm\)0.46} \\
        \bottomrule
        \arrayrulecolor{black}
        \midrule
        \multicolumn{11}{c}{\textbf{\textit{guided with IPR keywords}}} \\
        \midrule
        \arrayrulecolor{gray!20}
        Natural & 611.99 & 9.73 & 9.47 & 2.23 & 44.05 & 0.48 & 75.77 & 68.85 & 11.13 & 50.92 \\
        \midrule
        Random(U) & 2475.07\(\pm\)10.83 & 21.76\(\pm\)0.02 & 22.21\(\pm\)0.02 & 0.69\(\pm\)0.06 & 35.18\(\pm\)0.07 & 0.01\(\pm\)0.00 & 23.40\(\pm\)0.07 & 0.08\(\pm\)0.07 & 24.42\(\pm\)0.02 & 0.11\(\pm\)0.00 \\
        Random(E) & 3104.89\(\pm\)24.07 & 18.67\(\pm\)0.02 & 19.05\(\pm\)0.02 & 1.16\(\pm\)0.02 & 40.91\(\pm\)0.08 & 0.02\(\pm\)0.01 & 26.29\(\pm\)0.18 & 0.08\(\pm\)0.07 & 24.34\(\pm\)0.03 & 0.19\(\pm\)0.07 \\
        \midrule
        GPT-5 & 814.85\(\pm\)16.43 & 11.26\(\pm\)0.06 & 11.39\(\pm\)0.05 & 3.97\(\pm\)0.18 & 61.63\(\pm\)0.05 & 13.25\(\pm\)0.27 & 32.57\(\pm\)0.22 & 1.61\(\pm\)0.41 & 23.96\(\pm\)0.05 & 0.77\(\pm\)0.18 \\
        \midrule
        ESM3 & \second{330.44\(\pm\)9.90} & \second{6.33\(\pm\)0.07} & \second{6.59\(\pm\)0.07} & 28.13\(\pm\)0.24 & 68.98\(\pm\)0.42 & 21.11\(\pm\)0.47 & 60.90\(\pm\)0.77 & 32.93\(\pm\)2.43 & 16.73\(\pm\)0.27 & 22.68\(\pm\)1.83 \\
        CFP-Gen & \first{135.57\(\pm\)4.51} & \first{4.94\(\pm\)0.12} & \first{5.03\(\pm\)0.11} & 11.86\(\pm\)0.29 & 59.17\(\pm\)0.57 & 13.57\(\pm\)0.88 & \first{76.36\(\pm\)0.35} & \first{72.52\(\pm\)1.45} & \second{12.54\(\pm\)0.26} & \second{47.23\(\pm\)2.51} \\
        ProteinDT & 1506.64\(\pm\)5.70 & 11.87\(\pm\)0.02 & 11.93\(\pm\)0.02 & 10.02\(\pm\)0.38 & 65.68\(\pm\)0.30 & 5.83\(\pm\)0.39 & 37.59\(\pm\)0.15 & 0.04\(\pm\)0.07 & 26.19\(\pm\)0.03 & 0.00\(\pm\)0.00 \\
        Chroma & 1336.19\(\pm\)7.55 & 12.17\(\pm\)0.01 & 12.39\(\pm\)0.02 & \third{2.60\(\pm\)0.08} & \third{54.53\(\pm\)0.08} & \second{0.54\(\pm\)0.03} & 59.76\(\pm\)0.26 & 23.75\(\pm\)1.35 & 14.67\(\pm\)0.05 & 31.38\(\pm\)0.75 \\
        PAAG & 2748.12\(\pm\)25.25 & 17.85\(\pm\)0.04 & 18.06\(\pm\)0.03 & \second{2.32\(\pm\)0.11} & \second{44.78\(\pm\)0.06} & \first{0.08\(\pm\)0.01} & 30.89\(\pm\)0.03 & 0.11\(\pm\)0.11 & 24.98\(\pm\)0.02 & 0.19\(\pm\)0.13 \\
        Pinal & \third{525.38\(\pm\)80.49} & \third{8.12\(\pm\)0.45} & \third{8.22\(\pm\)0.47} & 16.73\(\pm\)1.96 & 59.97\(\pm\)3.55 & 6.32\(\pm\)1.28 & \third{65.69\(\pm\)5.42} & \third{44.90\(\pm\)12.43} & \third{14.10\(\pm\)2.19} & \third{36.13\(\pm\)9.57} \\
        ProDVa & 574.60\(\pm\)5.52 & 12.47\(\pm\)0.77 & 19.07\(\pm\)1.06 & \first{1.99\(\pm\)0.02} & \first{21.64\(\pm\)0.11} & \third{1.51\(\pm\)0.12} & \second{72.80\(\pm\)0.48} & \second{60.65\(\pm\)0.65} & \first{6.86\(\pm\)0.10} & \first{79.92\(\pm\)1.03} \\
        \bottomrule
        \arrayrulecolor{black}
        \midrule
        \multicolumn{11}{c}{\textbf{\textit{guided with IPR\&GO keywords}}} \\
        \midrule
        \arrayrulecolor{gray!20}
        Natural & 534.49 & 8.96 & 8.66 & 2.16 & 45.01 & 0.44 & 77.17 & 73.15 & 10.48 & 54.9 \\
        \midrule
        Random(U) & 2482.06\(\pm\)21.06 & 21.72\(\pm\)0.01 & 22.14\(\pm\)0.01 & 0.73\(\pm\)0.05 & 36.12\(\pm\)0.06 & 0.01\(\pm\)0.00 & 22.85\(\pm\)0.10 & 0.00\(\pm\)0.00 & 24.72\(\pm\)0.01 & 0.00\(\pm\)0.00 \\
        Random(E) & 3120.95\(\pm\)10.54 & 18.68\(\pm\)0.02 & 19.03\(\pm\)0.01 & 1.14\(\pm\)0.03 & 42.08\(\pm\)0.02 & 0.02\(\pm\)0.00 & 25.60\(\pm\)0.06 & 0.00\(\pm\)0.00 & 24.59\(\pm\)0.02 & 0.00\(\pm\)0.00 \\
        \midrule
        GPT-5 & 769.53\(\pm\)13.47 & 10.87\(\pm\)0.14 & 10.99\(\pm\)0.13 & 4.63\(\pm\)0.14 & 64.60\(\pm\)0.26 & 15.41\(\pm\)0.61 & 32.12\(\pm\)0.11 & 1.93\(\pm\)0.45 & 24.55\(\pm\)0.02 & 0.69\(\pm\)0.45 \\
        \midrule
        CFP-Gen & \first{163.51\(\pm\)6.62} & \first{5.23\(\pm\)0.04} & \first{5.23\(\pm\)0.07} & 13.14\(\pm\)1.18 & 59.86\(\pm\)0.53 & 14.17\(\pm\)0.32 & \second{72.70\(\pm\)1.07} & \second{60.90\(\pm\)1.11} & \third{14.45\(\pm\)0.28} & \second{42.69\(\pm\)1.76} \\
        ProteinDT & 1697.89\(\pm\)9.15 & 12.81\(\pm\)0.05 & 12.87\(\pm\)0.05 & 6.81\(\pm\)0.17 & 63.58\(\pm\)0.36 & \third{2.91\(\pm\)0.21} & 36.46\(\pm\)0.31 & 0.20\(\pm\)0.17 & 25.75\(\pm\)0.06 & 0.00\(\pm\)0.00 \\
        Chroma & 1360.90\(\pm\)5.58 & 12.19\(\pm\)0.04 & \third{12.40\(\pm\)0.04} & \second{2.53\(\pm\)0.12} & \third{55.65\(\pm\)0.16} & \second{0.56\(\pm\)0.05} & 58.71\(\pm\)0.59 & 19.29\(\pm\)2.67 & 15.33\(\pm\)0.30 & 29.72\(\pm\)1.34 \\
        PAAG & 2807.41\(\pm\)13.89 & 17.80\(\pm\)0.01 & 17.98\(\pm\)0.02 & \first{2.32\(\pm\)0.04} & \second{47.64\(\pm\)0.24} & \first{0.09\(\pm\)0.01} & 30.05\(\pm\)0.27 & 0.00\(\pm\)0.00 & 25.69\(\pm\)0.10 & 0.00\(\pm\)0.00 \\
        Pinal & \second{442.23\(\pm\)69.42} & \second{7.39\(\pm\)0.49} & \second{7.49\(\pm\)0.51} & 16.22\(\pm\)2.42 & 59.71\(\pm\)3.71 & 6.07\(\pm\)1.49 & \third{69.32\(\pm\)5.10} & \third{53.56\(\pm\)12.44} & \second{12.97\(\pm\)2.18} & \third{42.53\(\pm\)10.69} \\
        ProDVa & \third{500.40\(\pm\)7.93} & \third{10.48\(\pm\)0.07} & 13.61\(\pm\)0.95 & \third{2.61\(\pm\)0.43} & \first{28.73\(\pm\)0.63} & 3.89\(\pm\)0.68 & \first{74.26\(\pm\)0.27} & \first{67.46\(\pm\)1.49} & \first{8.06\(\pm\)0.07} & \first{72.16\(\pm\)0.23} \\
        \arrayrulecolor{black}
        \bottomrule
    \end{tabular}
    }
\end{table}

%% file: tabs/app/results/keyw/alignment.tex
\begin{table}[!htb]
    \caption{Benchmark results of Language Alignment for the keyword-guided task.}
    \label{tab:app-res-keyw-align}
    \centering
    \resizebox{\textwidth}{!}{
    \begin{tabular}{lcccccccccccccc}
        \toprule
        \multirow{2}{*}[-0.5ex]{Models} & \multicolumn{3}{c}{Model-based Alignment} & \multicolumn{9}{c}{Retrieval-based Alignment} \\
        \cmidrule[0.5pt](lr){2-4} \cmidrule[0.5pt](lr){5-13}
        & \makecell{ProTrek Score \(\uparrow\)} 
        & \makecell{IPR Recovery \(\uparrow\)}
        & \makecell{GO Recovery \(\uparrow\)}
        & \makecell{Soft(4) \(\uparrow\)} 
        & \makecell{Soft(10) \(\uparrow\)} 
        & \makecell{Soft(20) \(\uparrow\)} 
        & \makecell{Normal(4) \(\uparrow\)} 
        & \makecell{Normal(10) \(\uparrow\)} 
        & \makecell{Normal(20) \(\uparrow\)} 
        & \makecell{Hard(4) \(\uparrow\)} 
        & \makecell{Hard(10) \(\uparrow\)} 
        & \makecell{Hard(20) \(\uparrow\)} \\ 
        \midrule
        \multicolumn{13}{c}{\textbf{\textit{guided with GO keywords}}} \\
        \midrule
        \arrayrulecolor{gray!20}
        Natural & 21.6 & 100.0 & 100.0 & 94.52 & 92.78 & 89.75 & 87.59 & 77.49 & 69.41 & 37.23 & 28.72 & 26.7 \\
        \midrule
        Random(U) & 4.29\(\pm\)0.04 & 0.00\(\pm\)0.00 & 20.79\(\pm\)0.37 & 30.06\(\pm\)1.33 & 14.00\(\pm\)4.45 & 7.50\(\pm\)2.57 & 26.89\(\pm\)1.83 & 10.00\(\pm\)0.08 & 5.05\(\pm\)1.44 & 30.01\(\pm\)3.21 & 9.72\(\pm\)0.46 & 5.05\(\pm\)0.29 \\
        Random(E) & 3.44\(\pm\)0.01 & 0.00\(\pm\)0.00 & 11.71\(\pm\)1.48 & 29.87\(\pm\)0.50 & 12.41\(\pm\)2.18 & 6.69\(\pm\)1.04 & 27.08\(\pm\)0.30 & 11.06\(\pm\)0.55 & 5.96\(\pm\)0.98 & 28.62\(\pm\)2.22 & 11.98\(\pm\)0.66 & 5.87\(\pm\)0.79 \\
        \midrule
        GPT-5 & 6.85\(\pm\)0.17 & 3.76\(\pm\)0.23 & 31.34\(\pm\)1.09 & 71.52\(\pm\)3.43 & 59.55\(\pm\)2.09 & 48.68\(\pm\)1.31 & 60.22\(\pm\)1.34 & 38.05\(\pm\)1.67 & 26.31\(\pm\)0.74 & 28.76\(\pm\)2.62 & 12.70\(\pm\)2.29 & 7.94\(\pm\)1.52 \\
        \midrule
        ProteoGAN & 4.42\(\pm\)0.02 & 0.00\(\pm\)0.00 & 14.99\(\pm\)1.36 & 38.64\(\pm\)2.86 & 19.84\(\pm\)2.14 & 10.27\(\pm\)1.11 & 32.70\(\pm\)1.42 & 13.84\(\pm\)0.69 & 8.13\(\pm\)0.52 & 27.16\(\pm\)1.67 & 10.90\(\pm\)1.05 & 5.65\(\pm\)0.53 \\
        CFP-Gen & \third{10.03\(\pm\)0.33} & \third{9.67\(\pm\)1.07} & 18.98\(\pm\)1.14 & \third{66.24\(\pm\)2.17} & \third{57.40\(\pm\)1.85} & \third{47.39\(\pm\)3.10} & \third{53.46\(\pm\)2.49} & \third{38.87\(\pm\)2.00} & \third{30.09\(\pm\)0.92} & 27.80\(\pm\)3.72 & \third{13.37\(\pm\)0.74} & \third{8.15\(\pm\)0.42} \\
        ProteinDT & 1.70\(\pm\)0.21 & 0.03\(\pm\)0.05 & 18.52\(\pm\)0.71 & 37.52\(\pm\)6.22 & 16.59\(\pm\)1.25 & 8.90\(\pm\)0.36 & 34.05\(\pm\)1.91 & 15.39\(\pm\)1.42 & 8.51\(\pm\)0.88 & 27.08\(\pm\)1.79 & 12.51\(\pm\)1.23 & 6.30\(\pm\)0.73 \\
        Chroma & 1.84\(\pm\)0.03 & 0.23\(\pm\)0.05 & 16.33\(\pm\)2.36 & 32.13\(\pm\)0.79 & 13.80\(\pm\)2.03 & 7.74\(\pm\)0.87 & 27.90\(\pm\)0.96 & 12.07\(\pm\)1.06 & 5.92\(\pm\)0.38 & 27.99\(\pm\)0.43 & 12.22\(\pm\)0.92 & 5.15\(\pm\)0.82 \\
        PAAG & 4.38\(\pm\)0.17 & 0.00\(\pm\)0.00 & \third{21.66\(\pm\)2.71} & 38.58\(\pm\)1.75 & 20.49\(\pm\)1.59 & 11.16\(\pm\)1.42 & 34.10\(\pm\)0.92 & 16.45\(\pm\)0.76 & 8.85\(\pm\)0.79 & \third{31.36\(\pm\)2.33} & 12.07\(\pm\)0.46 & 6.30\(\pm\)1.21 \\
        Pinal & \second{12.69\(\pm\)1.42} & \second{19.26\(\pm\)1.90} & \second{22.76\(\pm\)1.78} & \second{73.98\(\pm\)4.37} & \second{61.90\(\pm\)8.29} & \first{56.66\(\pm\)7.96} & \second{61.52\(\pm\)4.86} & \second{49.93\(\pm\)3.82} & \second{42.95\(\pm\)3.35} & \second{35.88\(\pm\)4.96} & \second{21.26\(\pm\)2.48} & \second{17.89\(\pm\)2.27} \\
        ProDVa & \first{14.42\(\pm\)0.07} & \first{20.22\(\pm\)0.14} & \first{30.24\(\pm\)0.75} & \first{86.48\(\pm\)1.78} & \first{71.38\(\pm\)5.34} & \second{55.80\(\pm\)0.60} & \first{66.43\(\pm\)0.58} & \first{52.38\(\pm\)0.66} & \first{45.07\(\pm\)0.30} & \first{35.93\(\pm\)3.38} & \first{21.45\(\pm\)2.75} & \first{18.13\(\pm\)0.96} \\
        \bottomrule
        \arrayrulecolor{black}
        \midrule
        \multicolumn{13}{c}{\textbf{\textit{guided with IPR keywords}}} \\
        \midrule
        \arrayrulecolor{gray!20}
        Natural & 25.29 & 100.0 & 100.0 & 98.51 & 96.67 & 95.17 & 91.72 & 83.22 & 75.75 & 40.8 & 32.76 & 30.69 \\
        \midrule
        Random(U) & 7.53\(\pm\)0.07 & 0.00\(\pm\)0.00 & 25.75\(\pm\)3.21 & 27.70\(\pm\)3.13 & 10.84\(\pm\)1.76 & 6.44\(\pm\)0.75 & 26.44\(\pm\)1.78 & 11.72\(\pm\)0.11 & 6.21\(\pm\)0.64 & 28.05\(\pm\)1.44 & 9.85\(\pm\)0.63 & 4.87\(\pm\)0.24 \\
        Random(E) & 6.11\(\pm\)0.08 & 0.00\(\pm\)0.00 & 13.06\(\pm\)1.17 & 27.62\(\pm\)2.25 & 13.18\(\pm\)1.93 & 8.01\(\pm\)1.29 & 28.51\(\pm\)2.22 & 12.80\(\pm\)0.13 & 6.74\(\pm\)0.93 & 26.36\(\pm\)0.65 & 11.34\(\pm\)1.53 & 5.48\(\pm\)0.27 \\
        \midrule
        GPT-5 & 11.00\(\pm\)0.05 & 6.33\(\pm\)0.30 & 33.58\(\pm\)1.20 & 73.83\(\pm\)2.48 & 61.00\(\pm\)1.23 & 50.42\(\pm\)0.63 & 60.96\(\pm\)0.63 & 41.99\(\pm\)0.35 & 29.69\(\pm\)0.88 & 28.12\(\pm\)3.00 & 12.99\(\pm\)1.63 & 7.82\(\pm\)1.11 \\
        \midrule
        ESM3 & 6.22\(\pm\)0.18 & 20.17\(\pm\)0.86 & 15.43\(\pm\)2.74 & 55.41\(\pm\)7.66 & 37.77\(\pm\)1.88 & 31.64\(\pm\)1.72 & 48.31\(\pm\)1.52 & 33.01\(\pm\)1.24 & 26.47\(\pm\)1.63 & \third{29.82\(\pm\)0.88} & 14.69\(\pm\)0.18 & 10.69\(\pm\)0.43 \\
        CFP-Gen & \third{10.21\(\pm\)0.16} & \first{32.79\(\pm\)0.93} & \second{23.41\(\pm\)1.82} & \third{64.47\(\pm\)1.84} & \third{50.78\(\pm\)1.88} & \third{43.36\(\pm\)0.65} & \third{55.90\(\pm\)1.73} & \third{40.96\(\pm\)1.19} & \third{34.38\(\pm\)1.48} & 29.36\(\pm\)1.73 & \third{15.88\(\pm\)1.31} & \third{12.43\(\pm\)0.48} \\
        ProteinDT & 3.85\(\pm\)0.03 & 0.08\(\pm\)0.05 & \third{20.76\(\pm\)1.46} & 40.38\(\pm\)3.85 & 22.68\(\pm\)1.32 & 13.83\(\pm\)0.98 & 34.44\(\pm\)0.29 & 16.13\(\pm\)1.04 & 9.23\(\pm\)0.58 & 26.82\(\pm\)2.94 & 10.92\(\pm\)1.66 & 5.63\(\pm\)1.41 \\
        Chroma & 3.82\(\pm\)0.03 & 0.17\(\pm\)0.01 & 17.15\(\pm\)1.93 & 37.13\(\pm\)1.05 & 17.05\(\pm\)3.30 & 9.27\(\pm\)1.04 & 29.35\(\pm\)1.09 & 13.68\(\pm\)0.30 & 6.78\(\pm\)0.20 & 27.20\(\pm\)2.24 & 11.15\(\pm\)0.87 & 5.25\(\pm\)0.87 \\
        PAAG & 5.98\(\pm\)0.17 & 0.08\(\pm\)0.05 & 13.85\(\pm\)1.02 & 32.07\(\pm\)3.27 & 14.41\(\pm\)2.50 & 9.58\(\pm\)1.78 & 30.69\(\pm\)2.26 & 14.37\(\pm\)1.85 & 7.78\(\pm\)0.86 & 26.90\(\pm\)0.92 & 11.46\(\pm\)2.17 & 6.05\(\pm\)0.87 \\
        Pinal & \second{14.38\(\pm\)1.38} & \second{25.63\(\pm\)3.95} & 15.93\(\pm\)1.49 & \first{80.96\(\pm\)7.50} & \first{70.88\(\pm\)9.31} & \first{64.21\(\pm\)10.47} & \first{71.00\(\pm\)4.87} & \first{57.59\(\pm\)5.60} & \first{48.24\(\pm\)6.90} & \second{31.88\(\pm\)1.90} & \second{20.31\(\pm\)1.36} & \first{16.59\(\pm\)1.90} \\
        ProDVa & \first{15.19\(\pm\)0.19} & \third{24.58\(\pm\)0.69} & \first{26.59\(\pm\)0.40} & \second{80.15\(\pm\)2.25} & \second{64.90\(\pm\)3.11} & \second{57.05\(\pm\)4.15} & \second{65.29\(\pm\)0.64} & \second{51.99\(\pm\)0.93} & \second{44.44\(\pm\)0.88} & \first{33.26\(\pm\)1.24} & \first{20.73\(\pm\)1.50} & \second{15.90\(\pm\)0.54} \\
        \bottomrule
        \arrayrulecolor{black}
        \midrule
        \multicolumn{13}{c}{\textbf{\textit{guided with IPR\&GO keywords}}} \\
        \midrule
        \arrayrulecolor{gray!20}
        Natural & 27.36 & 100.0 & 100.0 & 99.55 & 99.11 & 98.96 & 93.62 & 87.39 & 79.82 & 45.85 & 36.65 & 34.27 \\
        \midrule
        Random(U) & 4.84\(\pm\)0.09 & 0.00\(\pm\)0.00 & 25.38\(\pm\)3.32 & 29.77\(\pm\)1.88 & 12.86\(\pm\)2.64 & 5.64\(\pm\)1.27 & 26.36\(\pm\)1.34 & 9.45\(\pm\)1.44 & 4.50\(\pm\)1.84 & 26.41\(\pm\)1.12 & 9.69\(\pm\)0.45 & 4.75\(\pm\)1.07 \\
        Random(E) & 3.72\(\pm\)0.09 & 0.00\(\pm\)0.00 & 14.67\(\pm\)4.14 & 30.42\(\pm\)2.20 & 12.12\(\pm\)0.62 & 5.93\(\pm\)0.90 & 27.00\(\pm\)1.46 & 11.28\(\pm\)1.04 & 6.28\(\pm\)1.30 & 26.56\(\pm\)1.46 & 10.29\(\pm\)1.30 & 5.24\(\pm\)0.31 \\
        \midrule
        GPT-5 & 9.13\(\pm\)0.10 & 6.64\(\pm\)0.31 & 29.43\(\pm\)1.02 & 77.50\(\pm\)2.08 & 63.60\(\pm\)2.31 & 54.35\(\pm\)1.57 & 61.08\(\pm\)2.41 & 42.78\(\pm\)2.27 & 29.97\(\pm\)1.51 & 26.61\(\pm\)0.45 & 12.81\(\pm\)0.67 & 7.52\(\pm\)0.17 \\
        \midrule
        CFP-Gen & \third{11.68\(\pm\)0.15} & \first{35.21\(\pm\)0.30} & \second{23.31\(\pm\)2.49} & \third{73.97\(\pm\)1.60} & \third{63.46\(\pm\)2.00} & \third{59.23\(\pm\)1.54} & \third{57.95\(\pm\)1.94} & \third{45.77\(\pm\)1.92} & \third{35.51\(\pm\)3.11} & \third{29.36\(\pm\)1.46} & \third{15.00\(\pm\)0.00} & \third{11.54\(\pm\)0.77} \\
        ProteinDT & 3.06\(\pm\)0.11 & 0.36\(\pm\)0.03 & 15.92\(\pm\)1.14 & 47.08\(\pm\)1.20 & 29.13\(\pm\)1.43 & 19.63\(\pm\)1.63 & 38.72\(\pm\)1.43 & 19.29\(\pm\)0.68 & 9.99\(\pm\)0.97 & 27.45\(\pm\)2.32 & 10.48\(\pm\)0.69 & 5.39\(\pm\)1.09 \\
        Chroma & 2.19\(\pm\)0.12 & 0.16\(\pm\)0.06 & 14.12\(\pm\)1.86 & 32.29\(\pm\)2.15 & 14.34\(\pm\)2.12 & 8.56\(\pm\)0.48 & 28.83\(\pm\)0.69 & 11.67\(\pm\)1.65 & 6.38\(\pm\)0.53 & 27.00\(\pm\)2.50 & 11.67\(\pm\)0.73 & 5.79\(\pm\)0.74 \\
        PAAG & 4.66\(\pm\)0.12 & 0.02\(\pm\)0.02 & 9.77\(\pm\)0.90 & 28.24\(\pm\)3.82 & 14.14\(\pm\)3.71 & 7.22\(\pm\)2.06 & 28.19\(\pm\)1.67 & 11.82\(\pm\)0.82 & 6.03\(\pm\)0.60 & 29.08\(\pm\)1.80 & 12.17\(\pm\)1.55 & 6.08\(\pm\)0.51 \\
        Pinal & \second{15.26\(\pm\)1.27} & \second{33.08\(\pm\)3.75} & \third{21.64\(\pm\)0.47} & \second{82.34\(\pm\)3.25} & \second{73.10\(\pm\)4.55} & \second{68.55\(\pm\)4.53} & \first{72.50\(\pm\)3.87} & \second{60.88\(\pm\)2.79} & \second{52.62\(\pm\)3.12} & \first{34.47\(\pm\)1.34} & \second{21.76\(\pm\)1.12} & \second{18.50\(\pm\)1.24} \\
        ProDVa & \first{16.78\(\pm\)0.12} & \third{30.95\(\pm\)0.56} & \first{25.24\(\pm\)0.45} & \first{82.54\(\pm\)1.19} & \first{74.18\(\pm\)1.36} & \first{69.24\(\pm\)1.63} & \second{71.51\(\pm\)0.93} & \first{61.23\(\pm\)1.50} & \first{52.97\(\pm\)2.43} & \second{33.88\(\pm\)1.79} & \first{22.45\(\pm\)1.09} & \first{19.93\(\pm\)1.09} \\
        \arrayrulecolor{black}
        \bottomrule
    \end{tabular}
    }
\end{table}

%% file: tabs/app/results/keyw/sim_nov_div.tex
\begin{table}[!htb]
    \caption{Benchmark results of Similarity, Novelty and Diversity for the keyword-guided task.}
    \label{tab:app-res-keyw-sim-nov-div}
    \centering
    \resizebox{\textwidth}{!}{
    \begin{tabular}{lccccccccccc}
        \toprule
        \multirow{2}{*}[-0.5ex]{Models} & \multicolumn{5}{c}{Similarity} & \multicolumn{4}{c}{Novelty} & \multicolumn{2}{c}{Diversity}\\
        \cmidrule[0.5pt](lr){2-6} \cmidrule[0.5pt](lr){7-10} \cmidrule[0.5pt](lr){11-12}
        & \makecell{GT-Identity \(\uparrow\)} 
        & \makecell{GT-TMScore \(\uparrow\)}
        & \makecell{ESM-F1 \(\uparrow\)} 
        & \makecell{ESM-Precision \(\uparrow\)} 
        & \makecell{ESM-Recall \(\uparrow\)} 
        & \makecell{Seq\textsubscript{Easy} \(\uparrow\)} 
        & \makecell{Seq\textsubscript{Hard} \(\uparrow\)} 
        & \makecell{Struct\textsubscript{Easy} \(\uparrow\)} 
        & \makecell{Struct\textsubscript{Hard} \(\uparrow\)} 
        & \makecell{Seq \(\uparrow\)} 
        & \makecell{Struct \(\uparrow\)} \\ 
        \midrule
        \multicolumn{12}{c}{\textbf{\textit{guided with GO keywords}}} \\
        \midrule
        \arrayrulecolor{gray!20}
        Natural & 100.0 & 100.0 & 100.0 & 100.0 & 100.0 & 44.34 & 4.07 & 56.96 & 18.15 & - & - \\
        \midrule
        Random(U) & 0.84\(\pm\)0.23 & 16.76\(\pm\)0.05 & 73.37\(\pm\)0.04 & 82.81\(\pm\)0.05 & 66.40\(\pm\)0.03 & 98.66\(\pm\)0.07 & 58.04\(\pm\)0.69 & 96.54\(\pm\)0.11 & 76.75\(\pm\)0.20 & 97.75 & 81.56 \\
        Random(E) & 0.63\(\pm\)0.16 & 17.03\(\pm\)0.03 & 74.24\(\pm\)0.04 & 83.70\(\pm\)0.05 & 67.24\(\pm\)0.03 & 98.44\(\pm\)0.04 & 60.28\(\pm\)0.09 & 95.85\(\pm\)0.15 & 75.92\(\pm\)0.19 & 99.35 & 81.54 \\
        \midrule
        GPT5 & 1.53\(\pm\)0.12 & 14.95\(\pm\)0.05 & 76.16\(\pm\)0.05 & 82.89\(\pm\)0.04 & 70.91\(\pm\)0.10 & 91.39\(\pm\)0.29 & 53.27\(\pm\)0.81 & 92.24\(\pm\)0.66 & 66.25\(\pm\)1.36 & 96.1 & 80.73 \\
        \midrule
        ProteoGAN & 0.28\(\pm\)0.07 & 14.75\(\pm\)0.24 & 74.25\(\pm\)0.07 & \first{84.37\(\pm\)0.10} & 66.84\(\pm\)0.05 & \second{99.13\(\pm\)0.06} & \second{65.24\(\pm\)0.27} & \second{96.19\(\pm\)0.17} & \first{75.82\(\pm\)0.29} & \second{99.31} & \second{84.37} \\
        CFP-Gen & \third{2.30\(\pm\)0.25} & 13.98\(\pm\)0.26 & 67.52\(\pm\)0.17 & 68.36\(\pm\)0.26 & 67.46\(\pm\)0.30 & 59.60\(\pm\)0.61 & 47.85\(\pm\)0.95 & 54.07\(\pm\)1.87 & 28.28\(\pm\)1.47 & 69.84 & \third{81.76} \\
        ProteinDT & 0.20\(\pm\)0.12 & 12.67\(\pm\)0.05 & \second{74.83\(\pm\)0.02} & \third{82.67\(\pm\)0.10} & 68.93\(\pm\)0.05 & \first{99.28\(\pm\)0.06} & \first{75.41\(\pm\)0.30} & \first{96.29\(\pm\)0.05} & \second{74.62\(\pm\)0.63} & \first{99.79} & \first{84.53} \\
        Chroma & 0.38\(\pm\)0.07 & \second{17.67\(\pm\)0.06} & \third{74.29\(\pm\)0.02} & 80.80\(\pm\)0.02 & \third{69.15\(\pm\)0.02} & 97.44\(\pm\)0.16 & 59.35\(\pm\)0.58 & 80.22\(\pm\)0.45 & 50.88\(\pm\)0.84 & 97.61 & 79.79 \\
        PAAG & 0.16\(\pm\)0.05 & \third{16.22\(\pm\)0.18} & \first{75.40\(\pm\)0.05} & \second{84.22\(\pm\)0.07} & 68.77\(\pm\)0.05 & \third{98.80\(\pm\)0.03} & \third{62.36\(\pm\)0.47} & \third{95.20\(\pm\)0.27} & \third{73.36\(\pm\)0.18} & \third{98.33} & 81.73 \\
        Pinal & \second{5.35\(\pm\)0.35} & 15.84\(\pm\)0.47 & 71.36\(\pm\)0.66 & 72.24\(\pm\)1.85 & \first{71.06\(\pm\)0.46} & 61.98\(\pm\)8.67 & 46.06\(\pm\)5.23 & 46.42\(\pm\)9.10 & 19.27\(\pm\)6.43 & 88.75 & 79.00 \\
        ProDVa & \first{9.07\(\pm\)0.14} & \first{20.25\(\pm\)0.13} & 72.54\(\pm\)0.10 & 75.81\(\pm\)0.06 & \second{70.25\(\pm\)0.12} & 48.37\(\pm\)0.72 & 25.02\(\pm\)0.81 & 62.12\(\pm\)0.19 & 32.72\(\pm\)0.02 & 42.84 & 32.02 \\
        \bottomrule
        \arrayrulecolor{black}
        \midrule
        \multicolumn{12}{c}{\textbf{\textit{guided with IPR keywords}}} \\
        \midrule
        \arrayrulecolor{gray!20}
        Natural & 100.0 & 100.0 & 100.0 & 100.0 & 100.0 & 44.92 & 4.47 & 59.46 & 20.09 & - & - \\
        \midrule
        Random(U) & 0.88\(\pm\)0.03 & 16.69\(\pm\)0.04 & 73.97\(\pm\)0.04 & 82.79\(\pm\)0.02 & 67.38\(\pm\)0.05 & 98.70\(\pm\)0.06 & 57.21\(\pm\)0.77 & 96.37\(\pm\)0.05 & 76.47\(\pm\)0.35 & 97.82 & 81.56 \\
        Random(E) & 0.70\(\pm\)0.15 & 16.85\(\pm\)0.12 & 74.79\(\pm\)0.03 & 83.65\(\pm\)0.03 & 68.16\(\pm\)0.04 & 98.44\(\pm\)0.08 & 59.57\(\pm\)0.81 & 95.62\(\pm\)0.14 & 75.05\(\pm\)0.48 & 99.35 & 81.46 \\
        \midrule
        GPT5 & 2.53\(\pm\)0.31 & 15.02\(\pm\)0.09 & 75.90\(\pm\)0.02 & 81.62\(\pm\)0.05 & 71.41\(\pm\)0.04 & 91.26\(\pm\)0.61 & 53.42\(\pm\)0.39 & 92.79\(\pm\)0.25 & 67.09\(\pm\)0.29 & 94.53 & 80.81 \\
        \midrule
        ESM3 & 4.43\(\pm\)0.22 & \first{21.30\(\pm\)0.28} & 72.22\(\pm\)0.09 & 75.06\(\pm\)0.13 & 69.96\(\pm\)0.25 & 85.30\(\pm\)0.68 & \second{71.87\(\pm\)1.23} & 73.80\(\pm\)1.08 & 37.56\(\pm\)0.41 & 93.69 & 76.79 \\
        CFP-Gen & \first{7.75\(\pm\)0.16} & 16.73\(\pm\)0.42 & 66.82\(\pm\)0.17 & 68.61\(\pm\)0.13 & 65.74\(\pm\)0.25 & 63.79\(\pm\)0.11 & 49.46\(\pm\)0.57 & 50.44\(\pm\)1.32 & 23.15\(\pm\)0.97 & 82.09 & \second{82.08} \\
        ProteinDT & 0.13\(\pm\)0.02 & 12.38\(\pm\)0.06 & \second{75.23\(\pm\)0.04} & \second{82.13\(\pm\)0.08} & \third{70.07\(\pm\)0.02} & \first{99.08\(\pm\)0.13} & \first{73.57\(\pm\)0.65} & \second{96.64\(\pm\)0.12} & \second{76.03\(\pm\)0.19} & \first{99.85} & \first{84.81} \\
        Chroma & 0.38\(\pm\)0.13 & \third{17.45\(\pm\)0.06} & \third{74.81\(\pm\)0.02} & \third{80.85\(\pm\)0.03} & 70.00\(\pm\)0.01 & \third{97.35\(\pm\)0.13} & 59.25\(\pm\)0.27 & \third{80.19\(\pm\)0.68} & \third{50.77\(\pm\)1.26} & \third{97.73} & 79.88 \\
        PAAG & 0.26\(\pm\)0.07 & 14.37\(\pm\)0.07 & \first{76.19\(\pm\)0.01} & \first{84.43\(\pm\)0.03} & 69.93\(\pm\)0.03 & \second{98.93\(\pm\)0.07} & \third{64.71\(\pm\)0.70} & \first{96.66\(\pm\)0.16} & \first{79.23\(\pm\)0.22} & \second{99.40} & \third{81.48} \\
        Pinal & \third{6.70\(\pm\)0.90} & 17.23\(\pm\)0.57 & 74.14\(\pm\)0.28 & 76.19\(\pm\)1.29 & \first{72.59\(\pm\)0.62} & 74.01\(\pm\)6.01 & 51.61\(\pm\)3.45 & 60.24\(\pm\)6.98 & 27.00\(\pm\)6.19 & 89.42 & 80.38 \\
        ProDVa & \second{7.39\(\pm\)0.08} & \second{20.75\(\pm\)0.36} & 73.31\(\pm\)0.09 & 76.62\(\pm\)0.17 & \second{70.99\(\pm\)0.09} & 51.80\(\pm\)1.34 & 28.86\(\pm\)1.37 & 65.58\(\pm\)0.77 & 31.26\(\pm\)1.57 & 50.11 & 43.44 \\
        \bottomrule
        \arrayrulecolor{black}
        \midrule
        \multicolumn{12}{c}{\textbf{\textit{guided with IPR\&GO keywords}}} \\
        \midrule
        \arrayrulecolor{gray!20}
        Natural & 100.0 & 100.0 & 100.0 & 100.0 & 100.0 & 43.23 & 3.89 & 56.17 & 17.72 & - & - \\
        \midrule
        Random(U) & 0.85\(\pm\)0.34 & 16.73\(\pm\)0.10 & 73.09\(\pm\)0.02 & 82.66\(\pm\)0.03 & 66.02\(\pm\)0.01 & 98.85\(\pm\)0.03 & 57.48\(\pm\)0.13 & 96.58\(\pm\)0.23 & 77.18\(\pm\)0.45 & 97.75 & 81.73 \\
        Random(E) & 0.69\(\pm\)0.04 & 16.89\(\pm\)0.05 & 73.97\(\pm\)0.06 & 83.59\(\pm\)0.05 & 66.85\(\pm\)0.05 & 98.54\(\pm\)0.03 & 60.06\(\pm\)0.40 & 95.79\(\pm\)0.02 & 75.98\(\pm\)0.28 & 99.29 & 81.57 \\
        \midrule
        GPT5 & 2.99\(\pm\)0.03 & 14.63\(\pm\)0.07 & 75.14\(\pm\)0.06 & 81.36\(\pm\)0.07 & 70.30\(\pm\)0.07 & 91.42\(\pm\)0.06 & 54.87\(\pm\)0.36 & 92.44\(\pm\)0.18 & 66.58\(\pm\)0.31 & 87.66 & 80.65 \\
        \midrule
        CFP-Gen & \third{8.00\(\pm\)0.30} & 16.12\(\pm\)0.13 & 66.07\(\pm\)0.27 & 68.60\(\pm\)0.12 & 64.41\(\pm\)0.40 & 65.78\(\pm\)1.27 & 54.72\(\pm\)1.90 & 52.97\(\pm\)1.96 & 28.89\(\pm\)0.77 & 75.16 & \second{81.91} \\
        ProteinDT & 0.29\(\pm\)0.14 & 13.08\(\pm\)0.03 & \second{74.91\(\pm\)0.02} & \second{82.88\(\pm\)0.06} & \third{68.91\(\pm\)0.06} & \second{98.75\(\pm\)0.06} & \first{71.44\(\pm\)0.50} & \second{96.53\(\pm\)0.05} & \second{75.73\(\pm\)0.21} & \first{99.67} & \first{84.18} \\
        Chroma & 0.29\(\pm\)0.08 & \third{17.51\(\pm\)0.10} & \third{74.09\(\pm\)0.06} & \third{80.76\(\pm\)0.07} & 68.84\(\pm\)0.05 & \third{97.49\(\pm\)0.07} & \third{59.36\(\pm\)0.05} & \third{80.76\(\pm\)0.50} & \third{51.45\(\pm\)1.07} & \third{97.85} & 79.97 \\
        PAAG & 0.20\(\pm\)0.08 & 13.77\(\pm\)0.07 & \first{75.52\(\pm\)0.05} & \first{84.45\(\pm\)0.05} & 68.81\(\pm\)0.05 & \first{99.12\(\pm\)0.06} & \second{65.07\(\pm\)0.68} & \first{97.22\(\pm\)0.09} & \first{81.53\(\pm\)0.18} & \second{99.42} & \third{81.51} \\
        Pinal & \second{9.41\(\pm\)1.27} & \second{18.11\(\pm\)0.94} & 74.00\(\pm\)0.38 & 75.46\(\pm\)1.50 & \first{73.01\(\pm\)0.68} & 66.95\(\pm\)6.76 & 49.03\(\pm\)4.31 & 54.41\(\pm\)7.66 & 22.43\(\pm\)6.35 & 85.67 & 78.20 \\
        ProDVa & \first{10.08\(\pm\)0.36} & \first{19.36\(\pm\)0.26} & 73.11\(\pm\)0.15 & 75.47\(\pm\)0.22 & \second{71.48\(\pm\)0.10} & 46.43\(\pm\)0.51 & 21.97\(\pm\)0.59 & 60.12\(\pm\)0.63 & 24.20\(\pm\)0.41 & 51.36 & 51.35 \\
        \arrayrulecolor{black}
        \bottomrule
    \end{tabular}
    }
\end{table}

%% file: docs/086LLM.tex
In addition to the aforementioned baselines, we aim to investigate whether large language models such as \textit{ChatGPT} can effectively perform the task. Using the API provided by \textit{OpenAI}, we accessed the GPT-5.1 model and employed it with a specialized system prompt and the function description as user input, as illustrated in Table~\ref{tab:set-gpt5}.

\begin{table}[!htb]
    \caption{Settings of GPT-5.1}
    \label{tab:set-gpt5}
    \centering
    \renewcommand{\arraystretch}{1.3}
    \resizebox{0.8\textwidth}{!}{
    \begin{tabular}{cl}
        \toprule
        \textbf{temperature} & 1.0\\
        \textbf{top\_p} & 0.1 \\
        \multirow{2}{*}{\textbf{system prompt}} & You are an expert in designing novel functional proteins tailored to user requirements. \\[-0.4ex]
        & Do not explain your answer; directly provide an amino acid sequence.\\
        \textbf{user prompt} & \(\{The\ functional\ description\ of\ the\ protein\}\) \\ 
        \bottomrule
    \end{tabular}
    }
\end{table}

We evaluated GPT-5.1 solely on keyword-guided tasks. As shown in Table~\ref{tab:app-res-keyw-sim-nov-div}\&\ref{tab:app-res-keyw-align}\&\ref{tab:app-res-keyw-sim-nov-div}, GPT-5.1 outperforms PAAG, Chroma, and ProteinDT across most language alignment metrics, though it still falls markedly short of CFP-Gen, Pinal, and ProDVa. It also achieves respectable results in both perplexity and repetitiveness, suggesting that large language models such as GPT-5.1 possess a solid grasp of protein sequences as well as textual data. However, GPT-5.1’s performance in Foldability ranks near the bottom among all models, revealing significant limitations in its understanding of protein structures. In terms of similarity, novelty, and diversity, GPT-5.1 performs on par with ProteinDT, Chroma, and PAAG, metrics that often do not directly indicate the quality of the designed sequences. 

%% file: docs/087Practice.tex
As noted in Section~\ref{sec:discuss-ppl-plddt}, the Perplexity, used to assess sequence plausibility, can to some extent capture two metrics of Foldability, the latter typically requiring extremely more computational time. In this section, we evaluate the computational costs of proposed metrics and propose a practical evaluation scheme.

Here, we randomly selected 50 and 500 samples from \keywtest and performed evaluations on a single machine equipped with 4 NVIDIA GeForce RTX3090 and one 48-core CPU. For Perplexity, we only evaluated the PPL-ProGen2; For Repetitiveness, we computed only the Repeat; For Foldability, both pLDDT and PAE metrics can be obtained simultaneously in a single inference of ESMFold; For Novelty and Diversity, we considered computing both Sequence and Structure metrics concurrently.

\input{tabs/app/extend/comp_time.tex}

As shown in Table~\ref{tab:comp-time}, the computational cost of Foldability is 160 times greater than that of Perplexity, as previously noted. The IPR Recovery is time-intensive, primarily because it relies on \textit{InterProScan}—a Java-based retrieval tool that cannot be accelerated using GPUs. In summary, we recommend using metrics other than Foldability and IPR Recovery for the initial screening, followed by these two metrics in a subsequent evaluation round. Furthermore, it should be noted that all metrics, except Repetitiveness, IPR Recovery, and GT-TMScore, depend on GPU processing. Therefore, the GPU requirements of \ourbench should not be underestimated.

%% file: tabs/app/extend/comp_time.tex
\begin{table}[!htb]
    \caption{Computation time for all metrics. The values below represent the mean time required to evaluate 50/500 samples.}
    \label{tab:comp-time}
    \centering
    \resizebox{\textwidth}{!}{
    \begin{tabular}{ccccccccccccc}
        \toprule
        \multirow{2}{*}{\textbf{Num of Samples}} &
        \multicolumn{2}{c}{\textbf{Plausibility}} & 
        \multirow{2}{*}{\textbf{Foldability}} & 
        \multicolumn{4}{c}{\textbf{Language Alignment}} &
        \multicolumn{3}{c}{\textbf{Similarity}} &
        \multirow{2}{*}{\textbf{Novelty}} &
        \multirow{2}{*}{\textbf{Diversity}} \\
        & Perplexity & Repetitiveness & & ProTrek Score & IPR Recovery & Retrieval Accuracy & GO Recovery & ESMScore & GT-Identity & GT-TMScore & & \\
        \midrule
        50 & 39s & 2s & 9m34s & 32s & 3s & 38s & 65s & 11s & 18s & 18s & 19s & 21s \\
        500 & 23s & 3s & 61m50s & 47s & 29m39s & 104s & 103s & 24s & 91s & 35s & 38s & 18s \\
        \bottomrule
    \end{tabular}
    }
\end{table}

%% file: docs/088Limitation.tex
In this study, we conducted a fair and comprehensive benchmark of  two tasks, 8 models, including ProteinDT, Chroma, PAAG, Pinal, ProDVa, ProteoGAN, ESM3, CFP-Gen across all \mnum metrics. For keyword-guided tasks, to ensure a fair comparison, we restricted the evaluation dataset to 1,057 proteins released between January 1, 2025, and August 25, 2025. It should be noted that the limited test data and the diversity of training datasets used by the baselines may affect the reliability of the evaluation results. Moreover, more than half of the baselines do not open-source their training data or code. As a result, achieving consensus on training procedures while evaluating only model architecture design remains unfeasible in this field. Therefore, PDFBench is dedicated to ensuring the quality of the test data to guarantee fair evaluation results across the publicly available model weights of each baseline to the greatest extent possible.